%% file: paper_final.tex
\documentclass[final]{cvpr}
\usepackage{multirow}
\usepackage{times}
\usepackage{epsfig}
\usepackage{graphicx}
\usepackage{graphbox}
\usepackage{amsmath}
\usepackage{amssymb}
\usepackage{bm}
\usepackage[flushleft]{threeparttable}
\usepackage{booktabs}
\usepackage{tabularx}

\usepackage[pagebackref=true,breaklinks=true,colorlinks,bookmarks=false]{hyperref}

\def\cvprPaperID{10157} 
\def\confYear{CVPR 2021}

\newcolumntype{Y}{>{\centering\arraybackslash}X}

\input{tex_shortcuts.tex}

\newcommand\blfootnote[1]{%
  \begingroup
  \renewcommand\thefootnote{}\footnote{#1}%
  \addtocounter{footnote}{-1}%
  \endgroup
}

\begin{document}

\title{Uncertainty-Aware Camera Pose Estimation from Points and Lines}

\author{Alexander Vakhitov$^{1}$\hspace{0.6cm}
Luis Ferraz Colomina$^{2}$ \hspace{0.6cm}
Antonio Agudo$^{3}$ \hspace{0.6cm}
Francesc  Moreno-Noguer$^{3}$ \hspace{0.6cm}\\
$^{1}$SLAMcore,  UK\\\
$^{2}$Kognia Sports Intelligence, Spain\\
$^{3}$Institut de Rob\`otica i Inform\`atica Industrial, CSIC-UPC, Spain
}


\maketitle

\begin{abstract}
\blfootnote{This work has been partially funded by the Spanish government under projects HuMoUR TIN2017-90086-R, ERA-Net Chistera project IPALM PCI2019-103386 and Mar\'ia de Maeztu Seal of Excellence MDM-2016-0656.} Perspective-n-Point-and-Line (P$n$PL) algorithms aim at fast, accurate, and robust camera localization with respect to a 3D model from 2D-3D feature correspondences, being a major part of modern robotic and AR/VR systems. 
Current point-based pose estimation methods use only 2D feature detection uncertainties, and the line-based methods do not take uncertainties into account. In our setup, both 3D  coordinates and 2D projections of the features are considered uncertain. We propose P$n$P(L)  solvers based on EP$n$P~\cite{lepetit2009} and DLS~\cite{hesch2011direct} for the uncertainty-aware pose estimation.
We also modify 
motion-only bundle adjustment to take 3D uncertainties into account. We perform exhaustive synthetic and real experiments on two different visual odometry datasets. The new P$n$P(L) methods outperform the state-of-the-art on real data in isolation, showing an increase in mean translation accuracy by 18\% on a representative subset of KITTI, while the new uncertain refinement improves pose accuracy for most of the solvers, e.g. decreasing mean translation error for the EP$n$P by 16\% compared to the standard refinement on the same dataset. The code is available at \href{https://alexandervakhitov.github.io/uncertain-pnp/}{https://alexandervakhitov.github.io/uncertain-pnp/}.

\end{abstract}

\input{intro.tex}

\input{related.tex}

\input{theory_cvpr.tex}

\input{experiments.tex}

{\small
\bibliographystyle{ieee}
\bibliography{egbib}
}

\end{document}

%% file: tex_shortcuts.tex
\newcommand{\bt}{\mathbf{t}}

\newcommand{\bht}{\mathbf{\hat t}}
\newcommand{\bx}{\mathbf{x}}

\newcommand{\bhx}{\mathbf{\hat x}}

\newcommand{\bff}{\mathbf{f}}

\newcommand{\bhf}{\mathbf{\hat{f}}}

\newcommand{\bu}{\mathbf{u}}

\newcommand{\Smat}{\mathtt{S}}
\newcommand{\bs}{\mathbf{s}}
\newcommand{\bc}{\mathbf{c}}

\newcommand{\bl}{\mathbf{l}}

\newcommand{\bp}{{\bf p}}
\newcommand{\bq}{{\bf q}}
\newcommand{\bP}{{\bf P}}
\newcommand{\bQ}{{\bf Q}}
\newcommand{\br}{{\bf r}}

\newcommand{\bw}{{\bf w}}
\newcommand{\bz}{{\bf z}}

\newcommand{\hR}{\mathtt{\hat R}}

\newcommand{\BSigma}{\mathbf{\Sigma}}

\newcommand{\btheta}{\bm{\theta}}

\newcommand{\balpha}{\bm{\alpha}}

\newcommand{\bnu}{\bm{\nu}}
\newcommand{\bpi}{\bm{\pi}}
\newcommand{\bbeta}{\bm{\eta}}
\newcommand{\by}{{\bf y}}

\newcommand{\R}{\mathtt{R}}
\newcommand{\I}{\mathtt{I}}

\newcommand{\erot}{{ {e}_{\textrm{rot}}}}

\newcommand{\etrel}{{e}_{\textrm{trans}}}

%% file: intro.tex
\section{Introduction}

Camera localization using sparse feature correspondences is a major part of augmented or virtual reality and robotic systems. The Perspective-$n$-Point(-and-Line), or P$n$P(L), methods can be successfully used to estimate the pose of a calibrated camera from sparse feature correspondences. Line features can increase localization accuracy in man-made self-similar environments which lack surfaces with distinctive textures~\cite{sola2012impact,zhang2015building,holzmann2016direct,pumarola2017pl}, motivating the use of P$n$P(L) methods~\cite{vakhitov2016accurate}.

While vision-based localization with respect to an automatically reconstructed map of sparse features is an important part of current robotic and AR/VR systems, commonly used P$n$P methods treat the features as absolutely accurate~\cite{zheng2013, hesch2011direct, lepetit2009, kneip2014upnp}. The more recent 2D covariance-aware methods relax this assumption~\cite{isprs-annals-III-3-131-2016,ferraz2014leveraging}, but only for the feature detections, still assuming perfect accuracy of the 3D feature coordinates.

\begin{figure}
\includegraphics[trim={10mm 0mm 0mm 0mm},clip,width=0.53\textwidth]{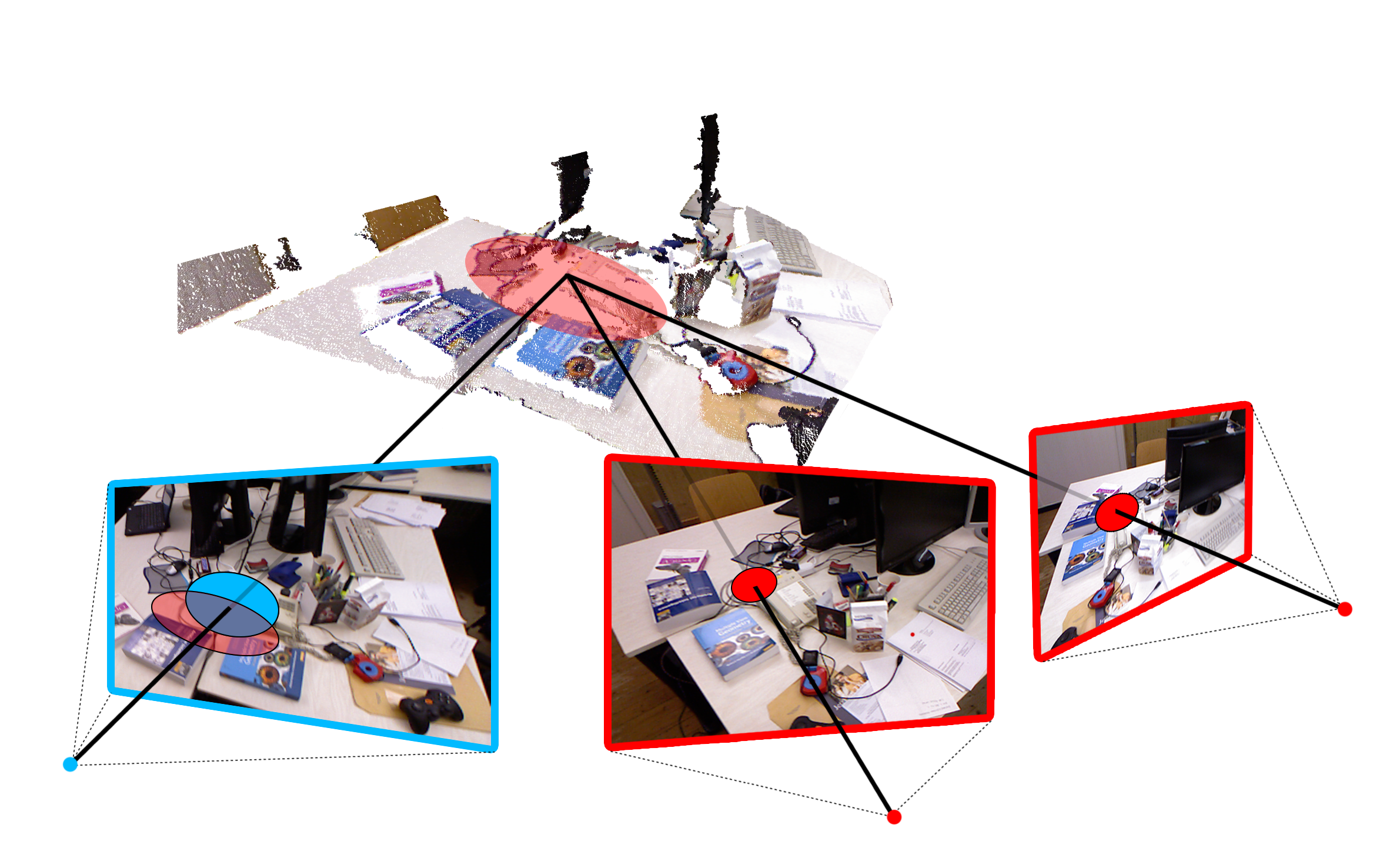}
\caption{We propose globally convergent P$n$P(L) solvers leveraging a complete set of 2D and 3D uncertainties for camera pose estimation. A 3D scene model with sparse features is reconstructed from images with known poses (right cameras), and we need to find a pose of a camera on the left. The point has 2D detection uncertainty (blue ellipsoid), and 3D model uncertainty  (red ellipsoid in the scene).}
\label{fig:teaser}
\vspace{-0.5cm}
\end{figure}
 In maps reconstructed with structure-from-motion, 3D feature coordinate accuracy can vary. Stereo triangulation errors grow quadratically with respect to the object-to-sensor distance, so the accuracy in estimating point depth can vary in several orders of magnitude, while the line stereo triangulation accuracy depends on the angle between the line direction and the baseline. Nevertheless, to the best of our knowledge, no prior method for P$n$P(L) was designed to take both 3D and 2D uncertainties into account. 

We propose to integrate the  feature uncertainty into the P$n$P(L) methods, see Fig.~\ref{fig:teaser}, which is our main contribution. We build on the classical DLS~\cite{hesch2011direct} and EP$n$P~\cite{lepetit2009} methods.  Additionally, we propose a modification to the standard 
nonlinear refinement, which is normally used after the P$n$P(L) solver, to take 3D uncertainties into account. 
An exhaustive evaluation on synthetic data and on the two real indoor and outdoor datasets demonstrates that new  P$n$P(L) methods are significantly more accurate than state-of-the-art,  both in isolation and in a complete pipeline, e.g. the proposed DLSU method reduces the mean translation error on KITTI by 18\%, see Section~\ref{sec:exp}. The proposed uncertain pose refinement can improve the pose accuracy by up to 16\% in exchange of an extra 5-10\% of the computational time. In a synthetic setting with noise in 2D feature detections, the new methods have the same accuracy as the most accurate 2D uncertainty-aware methods~\cite{ferraz2014leveraging,isprs-annals-III-3-131-2016}. The code is available at \href{https://alexandervakhitov.github.io/uncertain-pnp/}{https://alexandervakhitov.github.io/uncertain-pnp/}.

%% file: related.tex
\section{Related work}
We start with discussing how proposed methods relate to known P$n$P methods for arbitrary number of correspondences $n$, P$n$PL and 2D covariance-aware methods.
\vspace{-0.5cm}
\paragraph{Perspective-$n$-point.} Geometric gold standard cost~\cite{Ziss} for pose estimation for arbitrary $n$ is highly non-convex, and direct pose solvers rely on simplified {\em algebraic} costs. Early methods~\cite{oberkampf1996iterative,lu2000fast,schweighofer2008globally,quan1999linear,fiore2001efficient,ansar2003linear,abdel2015direct} were slow and inaccurate. 

Starting from~\cite{moreno2007accurate} fast direct solvers were developed~\cite{li2012robust,hesch2011direct,kneip2014upnp,zheng2013}. EP$n$P~\cite{moreno2007accurate} was the first to provide fast and accurate pose estimate solving a least-squares system with nonlinear constraints. 

EP$n$P was developed further in~\cite{lepetit2009,ferraz2014very,ferraz2014leveraging,vakhitov2016accurate}: \cite{lepetit2009} proposed adaptive PCA-based choice of control points and a fast iterative refinement step, \cite{ferraz2014very} designed an EPP$n$P method for robust pose estimation in presence outliers. Structure-from-motion~\cite{schonberger2016structure}, visual SLAM~\cite{mur2015orb}, object pose estimation~\cite{tekin2018real} rely on EP$n$P due to its robustness and fast computational time. A proposed solver EP$n$PU is a derivative of EP$n$P, providing better accuracy when feature uncertainty information is available.

Groebner basis solvers for polynomial systems are more accurate but are computationally more demanding~\cite{hesch2011direct,zheng2014general,zheng2013,kneip2014upnp,hadfield2019hard}. The DLS method~\cite{hesch2011direct} is fast enough for real-time use and has higher accuracy compared to the EP$n$P, while the most accurate method OP$n$P~\cite{zheng2013} is significantly slower.  
DLS minimizes the {\em object space error} and uses the Cayley parameterization, and has a singularity 
which can be avoided~\cite{nakano2015}. In this work, we build on the DLS and take feature uncertainties into account, however relying on the OP$n$P-like algebraic cost instead of the object space error.

\vspace{-0.5cm}
\paragraph{P$n$PL methods.} An early DLT method~\cite{Ziss} as well as an
 algorithm~\cite{ansar2003linear} can compute camera pose from $n$ line correspondences, but have inferior accuracy compared to new polynomial solver-based approaches~\cite{mirzaei2011globally,pribyl2015,zhang2012robust,kuang2014partial,Zhou19aaai}. Extending EP$n$P or OP$n$P to P$n$PL~\cite{vakhitov2016accurate,xu2017pose} is practical since a P$n$PL method uses all available mixed correspondences at once. In this work, we propose EP$n$PLU/DLSLU methods, which take line uncertainty into account in order to improve the pose estimates.
\vspace{-0.5cm}
\paragraph{P$n$P with uncertainty.} Features are detected with varying uncertainty, and 2D uncertainty-aware P$n$P methods~\cite{ferraz2014leveraging,isprs-annals-III-3-131-2016} use it to improve the estimated pose accuracy. CEPP$n$P~\cite{ferraz2014leveraging} builds on EPP$n$P and inherits the base method's low computational complexity. MLP$n$P~\cite{isprs-annals-III-3-131-2016}

 is more accurate and computationally demanding than CEPP$n$P, because it combines both a linear solver and a refinement into one method.
 
 Both CEPP$n$P and MLP$n$P use only 2D feature detection uncertainty and work only for points. 

In contrast, the approach we present in this paper is more accurate due to the use of both 2D and 3D feature uncertainty and works for a mixed set of line and point correspondences. 

%% file: theory_cvpr.tex
\section{Method}
We start with formulating the problem, and then proceed to introduce the uncertainty-aware pose solvers and the non-linear refinement method. We conclude the section with describing the approach for obtaining the feature covariances. We denote matrices, vectors and scalars with capital, bold and italic letters, e.g. $\R$, $\bx$, $\gamma$, and $x^{(i)}$ denotes the $i$-th component of $\bx$.
\subsection{PnP(L) with Uncertainty}

\begin{figure}[]
\centering 
\hspace{0.0cm}\includegraphics[trim={0 0 0 0 },clip,width=0.4\textwidth]{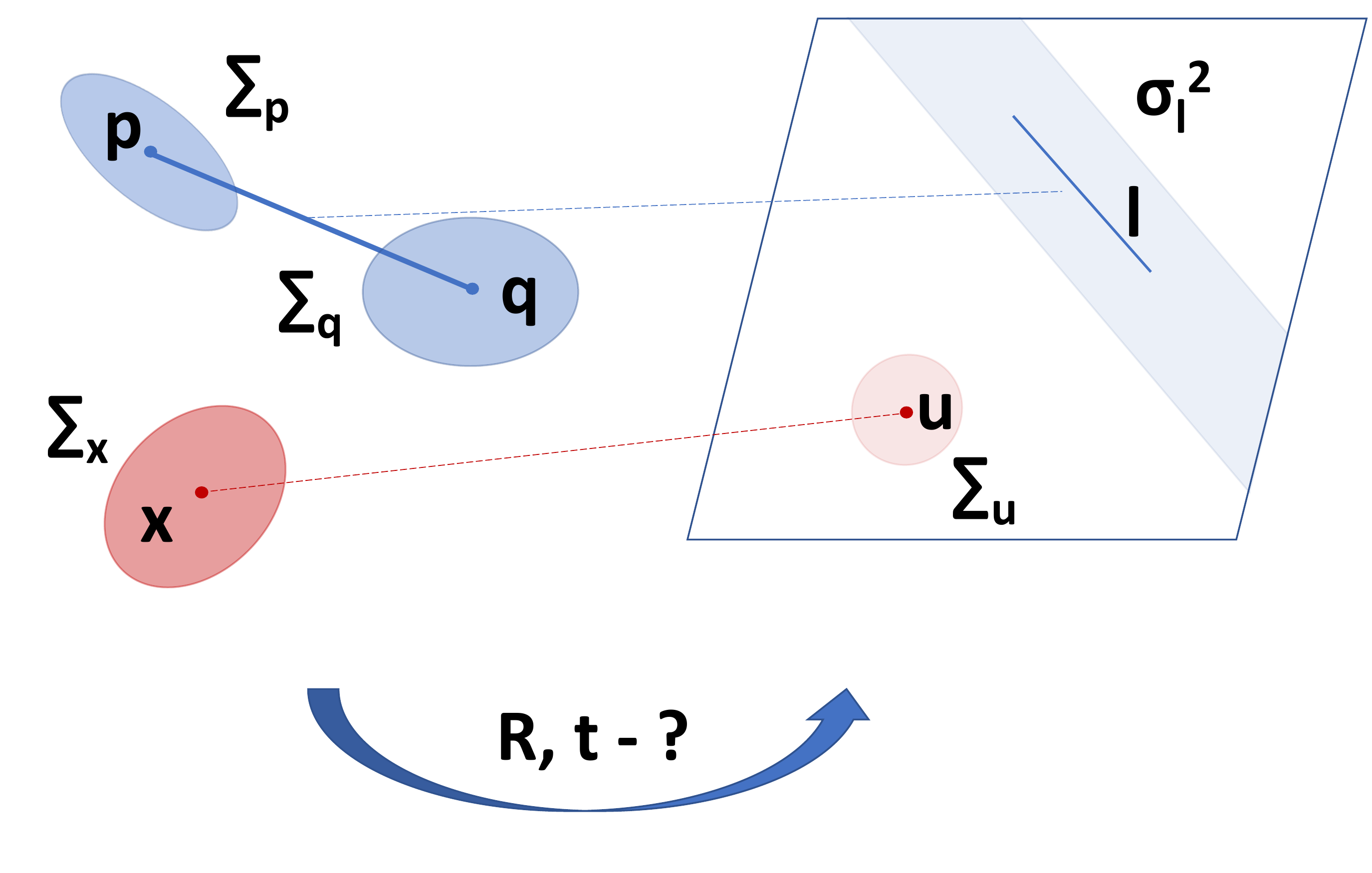}
  \caption{Schematic representation of the P$n$PL(U) problem, see text.}
  \label{fig:notation}
\vspace{-0.25cm}  
\end{figure}

We are given a set of $n_p$ 3D points $\{\bx_i\}_{i=1}^{n_p}$ and $n_l$ 3D line segments defined by their endpoints $\{\bp_i\}_{i=1}^{n_l}$, $\{\bq_i\}_{i=1}^{n_l}$. Points and line segment endpoints are corrupted by zero-mean Gaussian noises with covariances $\BSigma_{\bx_i}$, $\BSigma_{\bp_i}$ and $\BSigma_{\bq_i}$. The point projections  $\{\bu_i\}_{i=1}^{n_p}$ are corrupted with zero-mean Gaussian noises with covariances $\BSigma_{\bu_i}$. The line segments projections $\{\bl_i\}_{i=1}^{n_l}$ are represented as normalized line coefficients, so $\| \bl_i^{(1:2)}\| = 1$. We model the line segment detection uncertainty as a zero-mean Gaussian  added to the distance between the line and any point on the image plane, with a variance $\sigma_{l,i}^2$, see Fig.~\ref{fig:notation}. We consider a camera with known intrinsics, assuming that the camera calibration matrix $\mathtt{K}$ is an identity matrix.

Our problem is to estimate a rotation matrix $\mathtt{R}$ and a translation vector $\bt$ aligning the camera coordinate frame with the world frame. We assume knowledge of an  estimate of the average scene depth ${\bar d}$, and we consider also the case when there is a rough initial hypothesis $\hR, \bht$ available.

\subsection{Uncertainty for Pose Estimation Solvers}
In this section, we derive methods for uncertainty-aware pose estimation. We find the uncertainties for the algebraic feature projection residuals and incorporate them into the pose solvers in the form of the residual covariances. We start with point features, then move to lines.

\noindent{\bf Point residuals.} Let us parameterize a point in the camera coordinate frame as $\bhx(\btheta, \bx) = \R \bx + \bt,$ where $\btheta$ encodes the camera parameters $\R, \bt$. 
The algebraic point residual is based on perspective point projection:
\begin{equation}\label{eq:pt_residual}
    \br^{pt}(\btheta, \bx, \bu) = \bhx^{(1:2)}(\btheta, \bx) - \bu {\hat x}^{(3)}(\btheta, \bx),
\end{equation}
where $\bu$ is the projected point.

By our assumptions $\bx$ and $\bu$ are corrupted with additive zero-mean Gaussian noises with covariances $\BSigma_{\bx}$, $\BSigma_{\bu}$: $\bhx(\btheta, \bx) = \mathbb{E} \bhx + \bm{\xi}$ and ${\bu} = \mathbb{E} \bu + \bm{\zeta}$, where $\bm{\xi}$, $\bm{\zeta}$ are zero-mean Gaussian noise vectors. The covariance of ${\bhx}(\btheta, \bx)$ is:
\begin{equation}\label{eq:pt_cov_2d}
    \BSigma_{\bhx} = \R \BSigma_{\bx} \R^T = \left[ \begin{array}{cc} \Smat & \bw \\
    \bw^T & \gamma \end{array} \right].
\end{equation}
Substituting into (\ref{eq:pt_residual}), we obtain 
\begin{equation}\label{eq:pt_residual_dec}
     \br^{pt}=\mathbb{E} \{\bhx^{(1:2)} - \bu {\hat x}^{(3)} \}+ \bm{\xi}^{(1:2)} - \bu {\xi}^{(3)} - \bm{\zeta}{\hat x}^{(3)} - \xi^{(3)}\bm{\zeta},
\end{equation}
omitting the function arguments for clarity. By expressing the covariance of (\ref{eq:pt_residual_dec}) as $\mathbb{E} \br^{pt} (\br^{pt})^T$, using the independence of $\bm{\xi}$ and $\bm{\zeta}$ we obtain the residual covariance:
\begin{equation}\label{eq:pt_res_cov}
    \BSigma_{\br^{pt}} = 
    \Smat + \gamma \bu \bu^T + ({\hat x}^{(3)})^2 \BSigma_{\bu} - (\bu \bw^T + \bw \bu^T).
\end{equation}
To compute $\BSigma_{\br^{pt}}$, we need to know $\R$ 
and ${\hat x}^{(3)}$. If we approximate the model point covariance $\BSigma_{\bx} \approx \sigma^2 \I$, where $\I$ denotes the identity, then $\BSigma_{\bhx} \approx \sigma^2 \I$ as well, as follows from (\ref{eq:pt_cov_2d}). If we have a rough pose hypothesis $\hR, \bht$, we can use it instead to approximate $\BSigma_{\br^{pt}}$. We propose the new solvers in two modifications. In the first case, the solver uses the average scene depth estimate ${\bar d}$ to approximate the point depths, and an isotropic approximation to 3D point covariance (see Section~\ref{sec:cov} below), we dub these solvers EP$n$PU and DLSU. 
In the second case, the solver uses the pose hypothesis, typically available as an output of the RANSAC loop, to approximate the point depths and compute the 3D point covariance estimates; we denote these methods DLSU*, EP$n$PU*. 

\vspace{1mm}
\noindent{\bf Line residuals.} We are given the normalized 2D line segment coefficients $\bl$ as well as the segment 3D endpoints  $\bp, \bq$, and consider the algebraic line residual following~\cite{vakhitov2016accurate}:
\begin{equation}\label{eq:ln_residual}
    \br^{ln}(\btheta, \bp, \bq, \bl) = \left[
    \begin{array}{c}
    \bl^T {\hat \bp} (\btheta, \bp)\\
    \bl^T {\hat \bq}(\btheta, \bq)
    \end{array}\right],
\end{equation}
where ${\hat \bp}(\btheta, \bp),{\hat \bq}(\btheta, \bq)$ are the 3D endpoints in the camera coordinate frame, $ln$ stands for 'line'. Let us decompose the endpoint ${\hat \bp}(\btheta, \bp) = \mathbb{E} {\hat \bp}(\btheta, \bp) + \bbeta_{\bp}$, where $\bbeta_{\bp}$ has the covariance  $\BSigma_{{\hat \bp}} = \mathtt{R} \BSigma_{\bp} \mathtt{R}^T$. Under our model for the line detection noise, the noise-corrupted signed line-point distance is $\bl^T \by_h = \mathbb{E} \{ \bl^T\} \by_h + \bnu_y,$ where $\bnu_y$ is a zero-mean Gaussian with variance $\sigma_{l}^2$, $\by_h$ is  an arbitrary image point, in homogeneous coordinates. If $\by_h$ is a projection of a point $\by = \lambda_{\by} \by_h$ with depth $\lambda_{\by}$, then $\bl^T \by = \mathbb{E} \{ \bl^T \by \} + \lambda_{\by}\bnu_y$. Therefore,
\begin{equation}
\bl^T {\hat \bp}(\btheta, \bp) = \mathbb{E} \{\bl^T {\hat \bp}(\btheta, \bp) \} + \lambda_{\bp} \bnu_{\bp} + \bl_i^T \bbeta_{\bP_i},
\end{equation}
where $\bnu_{\bP_i}$ is the line detection noise. The variance is
\begin{equation}
    \mathbb{E} (\bl^T {\hat \bp} - \mathbb{E} \bl^T {\hat \bp})^2 = \lambda_{\bp}^2 \sigma_{l}^2 + \bl^T \BSigma_{{\hat \bp}} \bl,
\end{equation}
where we omit the function arguments for brewity. Under our model, the noise in $\bp$, $\bq$ and the line detection noises are assumed independent. We acknowledge that this is a simplification, however it speeds up the computations, and works in practice, as we show in the experiments. Moreover, other works, e.g.~\cite{pumarola2017pl}, rely on such a model as well, while in the offline setting one could follow~\cite{forstner2016photogrammetric} in using more advanced noise models for the lines. The covariance for the line residual is
\begin{equation}
    \BSigma_{\br^{ln}} = \sigma_{l}^2 \textrm{diag}(\lambda_{\bp}^2, \lambda_{\bq}^2) + \textrm{diag}(\bl^T \BSigma_{{\hat \bp}} \bl, \bl^T \BSigma_{{\hat \bq}} \bl).
\end{equation}
In order to compute the covariance we need $\mathtt{R}$ and the point depths $\lambda_{\bp}, \lambda_{\bq}$. For EP$n$PLU and DLSLU, we approximate the point depths with ${\bar d}$ given in the problem formulation and the covariances $\BSigma_{{\hat \bP_i}}$, $\BSigma_{{\hat \bQ_i}}$ as isotropic, for EP$n$PLU* and DLSLU* we use a pose hypothesis $\hR, \bht$ to compute these values.

So far we obtained a general form of residual covariances for point and line features under our noise model. Next we show, how to use it in the P$n$P(L) solvers.

\subsection{EPnP with Uncertainty}
We generalize the EP$n$P~\cite{lepetit2009} and EP$n$PL~\cite{vakhitov2016accurate}  to leverage 2D and 3D uncertainties in pose prediction. 
EP$n$P starts with computing an SE3-invariant barycentric representation $\balpha$ of a point $\bx$:
\begin{equation}\label{eq:epnp_proj}
    \bx = \mathtt{C} \balpha,
\end{equation}
where $\mathtt{C} = [\bc_1,\ldots,\bc_4]$ is a matrix of the four specifically chosen {\it control points} in the world coordinate frame. \cite{lepetit2009}~proposed to choose the first point $\bc_1$ as a mean of $\bx_i$ and $\bc_2,\bc_3$ and $\bc_4$ as the maximum variance directions computed using principal component analysis (PCA). Preliminary experiments show, that when 3D noise is added to $\bx$, the accuracy of the PCA version degrades. This motivated us to modify the control points choice to use the 3D uncertainties. We can get a straightworward theoretically solid PCA generalization under isotropic approximation of 3D point covariances $\BSigma_{\bx} \approx \sigma_{\bx}^2 \I,$ where $\sigma_{\bx}^2 = \frac{1}{3} \textrm{trace}(\BSigma_{\bx})$. In particular,  classical PCA solves a following problem to obtain the $j$-th principal direction:
\begin{equation}
    \sum_{i} (\bz^T {\tilde \bx_i})^2 \to \max_{\bz} \; \textrm{s.t.} \; \| \bz \| = 1,
\end{equation}
where ${\tilde \bx_i}$ are the centered points with subtracted projections on the first $j-1$ components, and $\bz$ is the sought principal direction. The covariance is $\textrm{cov}(\bz^T {\tilde \bx}) = \bz^T {\tilde \bx} {\tilde \bx}^T \bz = \bz^T \BSigma_{\tilde \bx} \bz = \sigma_{\bx}^2$ using the fact that $\| \bz \| = 1$. Then, we modify the problem as
\begin{equation}
    \sum_{i} \sigma_{\bx,i}^{-2} (\bz^T {\tilde \bx_i})^2 \to \max_{\bz} \; \textrm{s.t.} \; \| \bz \| = 1,
\end{equation}
see more details in the supp. mat.

The camera pose in EP$n$PL is represented through the control points in the camera coordinate frame, so $\btheta_{EPnP} = \mathtt{\hat C} = [{\hat \bc}_1,\ldots,{\hat \bc}_4],$ and ${\hat \bc}_i = \mathtt{R} \bc_i + \bt$. The camera frame point is  ${\hat \bx}(\btheta_{EPnP}, \bx) = \mathtt{\hat C} \balpha(\bx)$.
 EP$n$P(L) uses the algebraic residuals for lines and points (\ref{eq:pt_residual}, \ref{eq:ln_residual}), solving a problem 
  \begin{equation}
      \|\mathtt{M} 
      \textrm{vec} 
      (
      \mathtt{\hat C}
      )
      \|^2 \to \min_{\mathtt{\hat C}},
  \end{equation}
  where $\textrm{vec}(\cdot)$ denotes a vectorized matrix. The solution is given by an eigendecomposition of a $12\times12$  matrix $\mathtt{M}^T \mathtt{M}$. The method then looks for $\mathtt{\hat C}$ in the subspace of the eigenvectors of $\mathtt{M}^T \mathtt{M}$ with smallest eigenvalues.
 
 The proposed EP$n$P(L)U method follows the same strategy, constructing the uncertainty-augmented matrix $\mathtt{M}_U$. In the previous section we noted, that to estimate the covariances of the point and line algebraic residuals (\ref{eq:pt_residual}, \ref{eq:ln_residual}) we need to know $\mathtt{R}$ and $\bt$. However, approximating the point covariances by isotropic covariance matrices as $\BSigma_{\bx} \approx \sigma_{\bx}^2 \I$, $\BSigma_{\bp} \approx \sigma_{\bp}^2 \I$, $\BSigma_{\bq} \approx \sigma_{\bq}^2 \I$, where $\sigma_{\cdot}^2 = \frac{1}{3}\textrm{trace}(\BSigma_{\cdot})$, we get rid of the dependency on $\mathtt{R}$ in the residuals. By replacing the point depth with the approximate scene depth ${\bar d}$, we compute the point (\ref{eq:pt_residual}) and line (\ref{eq:ln_residual}) residual covariances  as
 \begin{equation}\label{eq:epnp_pt_residual}
     \BSigma_{\br^{pt}}^{EPnP} = \sigma_{\bx}^2 \I + {\bar d}^2 \BSigma_{\bu} + \sigma_{\bx}^2 \bu \bu^T.
 \end{equation}
\begin{equation}\label{eq:epnp_ln_residual}
     \BSigma_{\br^{ln}}^{EPnP} = \sigma_{l}^2 {\bar d}^2 \I + \|\bl\|^2 \textrm{diag} (\sigma_{\bp}^2, \sigma_{\bq}^2).
 \end{equation}
We also consider a case when a rough pose hypothesis is given. In this case, we still use the isotropic approximation of uncertainties, but use the pose to compute estimates of the depths of points. The method proceeds as the basis version. Next we describe our DLS-based approach.
\subsection{DLS with Uncertainty}
The DLS method~\cite{hesch2011direct} employs Cayley rotation parameterization to solve a least-squares polynomial system of the algebraic residuals for the point correspondences with the Groebner basis techniques. It relies on so-called {\em object space error} P$n$P, when one minimizes the distance between the backprojection ray of the point detection and the 3D point. However, we decided to use the algebraic residual (\ref{eq:pt_residual}), which allows for faster computations  and results in a method with similar accuracy, see supp.mat. for the comparison. DLS performs eigendecomposition of a $27 \times 27$ matrix. We keep the Cayley parameterization of DLS, but reformulate the equations, and generate the new solver of the same dimension using the generator of~\cite{larsson2017efficient}. 

The DLS uses the following parameterization of a point in camera coordinates: ${\hat \bx} (\btheta, \bx)= \mathtt{R}(\bs) \bx + \bt$, so $\btheta_{DLS} = [\bs^T, \bt]^T$, where $\bs \in \mathbb{R}^3$ is a vector of the Cayley rotation parameters: 
\begin{equation}
    \mathtt{R}(\bs) = \frac{1}{1+\|\bs\|^2}\left((1-\bs \bs^T) \I + 2 [\bs]_x + 2 \bs \bs^T\right),
\end{equation}
 $[\bs]_x$ denotes a cross product matrix. We use the residuals for lines and points (\ref{eq:ln_residual},\ref{eq:pt_residual}) with the camera parameterization $\btheta_{DLS}$. The residual covariances are obtained as in a case of EP$n$P. The point or line residual $\br_k(\bs, \bt)$ can be expressed using the DLS parameterization as 
\begin{equation}
    \br_k(\bs, \bt) = \mathtt{A}_k \textrm{vec}(\mathtt{R}(\bs)) + \mathtt{T}_k \bt,
\end{equation}
and we denote its covariance as $\BSigma_{\br_k}$.
The cost function of the method is
\begin{equation}\label{eq:dls_cost}
    \frac{1}{2}\sum_{k=1}^{n_r} \br_k^T(\bs, \bt) \BSigma_{\br_k}^{-1} \br_k(\bs, \bt) \to \min_{\bs, \bt}.
\end{equation}

We constrain the gradient of the cost by $\bt$ to be zero,  express $\bt$ using the remaining unknowns and obtain the cost that depends only on $\mathtt{R}(\bs)$. Following DLS, we multiply this cost by $(1+\|\bs\|^2)^2$, so it becomes a polynomial of $\bs$. We constrain its gradient by $\bs$ to be zero and obtain a third order polynomial system with three unknowns. It is solved using the generated solver, then $\bt$ is found using an expression obtained before, see the details of the derivation in the supp. mat. 

To improve accuracy, we also use an optional non-linear refinement stage by refining the cost (\ref{eq:dls_cost}) with a Newton method starting from the output of a solver, computing the Hessian of the cost analytically. 

One often refines the output of the pose solver with non-linear minimization of the gold-standard feature reprojection errors~\cite{Ziss}. In the following section we propose a new formulation of a refinement method in order to take the full set of feature uncertainties into account.

\subsection{Uncertainty-aware Pose Refinement}\label{sec:ref}

 When the structure is fixed, to obtain optimal estimates of the camera pose one uses {\em motion-only}  bundle adjustment~\cite{triggs1999bundle}, that is formulated as a non-linear least squares-based log-likelihood maximization. In feature-based pose estimation, one runs it as a final refinement step, initializing with the output of a pose solver. A standard 2D covariance-aware formulation of the {\em motion-only} bundle adjustment cost is:
 \begin{equation}\label{eq:ba_loss}
     {\cal L}(\btheta) = \sum_{i=1}^{n_p} \| {\bar \br}_i^{pt} \|_{\BSigma_{{\bar \br}_i^{pt}}}^2 + 
     \sum_{i=1}^{n_l} \| {\bar \br}_i^{ln} \|_{\BSigma_{{\bar \br}_i^{ln}}}^2 \to \min{\btheta},
 \end{equation}
 where $\btheta$ is the camera pose, ${\bar \br}_i^{pt}$, ${\bar \br}_i^{ln}$ are the gold standard point and line feature residuals, and $\| \cdot \|_{\BSigma}$ denotes the Mahalanobis distance with covariance $\BSigma$.
 The 'gold standard' residual for a point~\cite{Ziss} is
\begin{equation}\label{eq:ransac_pt_residual}
    {\bar \br}^{pt}(\bx, \mathtt{R}, \bt) = \bu - \bpi(\mathtt{R}\bx+ \bt),
\end{equation}
where the projection function $\bpi$ is defined as $\bpi({\hat \bx}) := \frac{1}{\hat x^{(3)}} \bx^{(1:2)}$, and ${\hat \bx}  = \mathtt{R} \bx + \bt$. A 'gold standard' residual for a line is
\begin{equation}\label{eq:ransac_ln_residual}
    {\bar \br}^{ln}(\bp, \bq, \mathtt{R}, \bt) = \left[ \begin{array}{c}
    \bl^T \bpi(\mathtt{R} \bp + \bt) \\
    \bl^T \bpi(\mathtt{R} \bq+\bt)
    \end{array}
    \right].
\end{equation}
 
 Visual odometry systems, e.g.~\cite{mur2015orb,mur2017orb}, use the 2D covariance of the feature detection as the residual covariance. This corresponds to setting $\BSigma_{{\bar \br}^{pt}} = \BSigma_{\bu}$, $\BSigma_{{\bar \br}^{ln}} = \sigma_{l}^2 \I$, and we will dub this scheme as {\em standard} refinement. It is often implemented based on a fast and efficient Levenberg-Marquardt method.
 
 In our case, we wish to use the full residual covariance, including both 2D and 3D uncertainty. For the point residual it is
 \begin{equation}\label{eq:ransac_pt_cov}
    \BSigma_{{\bar \br}^{pt}} = \BSigma_{\bu} + \mathtt{J}({\hat \bx}) \R \BSigma_{\bx} \R^T \mathtt{J}^T({\hat \bx}),
\end{equation}
where $\mathtt{J}$ is a Jacobian of $\bpi$ with respect to ${\hat \bx}$; for the line residual the covariance is 
\begin{equation}\label{eq:ransac_ln_cov}
\BSigma_{{\bar \br}^{ln}} = \sigma_{l}^2 \I + \textrm{diag}\left( \bl^T \BSigma_{\hat \bp}^{\bpi} \bl, \; \bl^T \BSigma_{\hat \bq}^{\bpi} \bl \right),
\end{equation}
where $\BSigma_{\bhf}^{\bpi} = \mathtt{J}(\bhf) \R \BSigma_{\bff}\R^T \mathtt{J}^T({\bhf})$,
for $\bhf = \{{\hat \bp}, {\hat \bq}\}$, $\bff = \{ \bp, \bq \}$.
 
 The covariances in the form (\ref{eq:ransac_pt_cov}, \ref{eq:ransac_ln_cov}) are not constant with respect to the camera pose. They cannot be used in a classical Gauss-Newton scheme.The cost (\ref{eq:ba_loss}) can be minimized using non-linear minimization, defining a {\em full uncertain} refinement method.
 
 However, the {\em full uncertain} refinement has a downside of being computationally inefficient compared to a {\em standard} refinement. Therefore, we propose a technique resembling Iterative Reweighted Least Squares, in which we make Gauss-Newton iterations, but update the estimate of the covariances~(\ref{eq:ransac_pt_cov}, \ref{eq:ransac_ln_cov}) on each step. We call it {\em (iterative) uncertain} refinement. This technique results in similar accuracy to the {\em full uncertain} refinement, but in terms of computational efficiency is comparable to the {\em standard} refinement, as we show in the experiments section below. In the following section, we explain our approach to obtaining the point and line uncertainties.

\subsection{Obtaining the Uncertainties}\label{sec:cov}
The 2D feature uncertainties can be obtained from a feature detector, e.g. for a multiscale pyramidal detector with a scale step of $\kappa$ we estimate  $\BSigma_{\bu} = \sigma_{o}^2\mathtt{I},$  $\sigma_{l}^2 = \sigma_{o}^2,$ where $\sigma_o = \kappa^{o-1}\epsilon$, and $o$ is a level of the image pyramid to which the feature belongs, $\epsilon$ is the feature detection accuracy. 

Uncertainty for the 3D point can be estimated after the triangulation following a standard error propagation technique, e.g.~\cite{Ziss}, Chapter 5. While there exists a single natural 3D point parameterization, the situation with line features is less clear. The line covariance formulation depends on the representation used for line triangulation, and there are several known parameterizations of lines,see~\cite{bartoli2005structure,zhao2018good,pumarola2017pl}. As long as we represent a line in 3D through the endpoints of some 3D line segment, we require a method to accurately find these endpoints and their covariances. We reconstruct the endpoints as the unknowns, and for the first camera we use the point-based reprojection residuals, while for the other cameras we use the line-based residuals, see the supp. mat. This way, we can use error propagation to obtain the line endpoint uncertainties after triangulating the line.

If we have an arbitrary positive semi-definite covariance $\BSigma_{\bx}$ of a 3D point, an isotropic approximation for it would be $\sigma_{\bx}^2\I,$ where $\sigma_{\bx}^2 = \frac{1}{3}\textrm{trace}(\BSigma_{\bx}).$ This approximation is optimal in the Frobenius norm sense: $\| \BSigma_{\bx} - \sigma_{\bx}^2\I\|_{F}^2 = \sum_{i=1}^3(\rho_i^2-\sigma_{\bx}^2)^2$, where $\rho_i^2$ are the singular values of $\BSigma_{\bx}$.

We have described the new methods and now continue with evaluating them in a synthetic and real settings.

%% file: experiments.tex
\begin{figure*}[t!]
\centering
\hspace{0cm}
\includegraphics[trim={5mm 5mm 88mm 15mm},clip,width=0.02\textwidth,align=c]{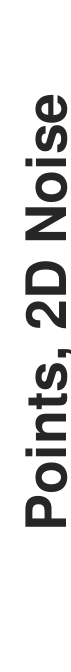} 
\includegraphics[trim={5mm 5mm 5mm 5mm},clip,width=0.24\textwidth,align=c]{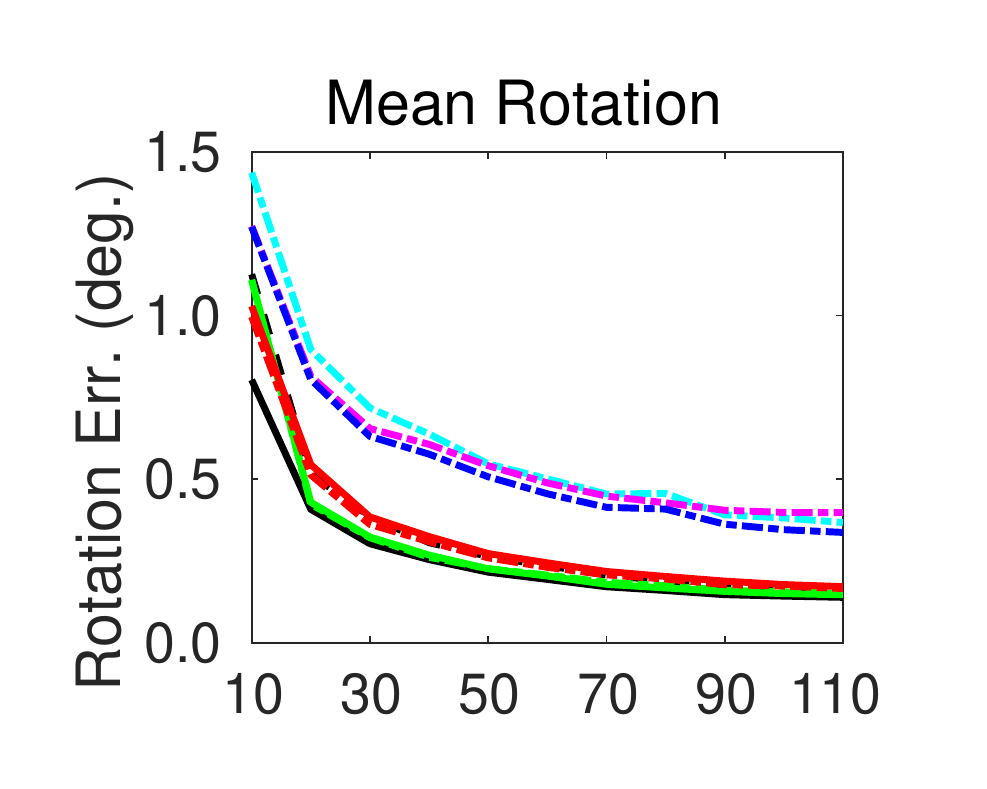} 
\hspace{0cm}\includegraphics[trim={5mm 5mm 5mm 5mm},clip,width=0.22\textwidth,align=c]{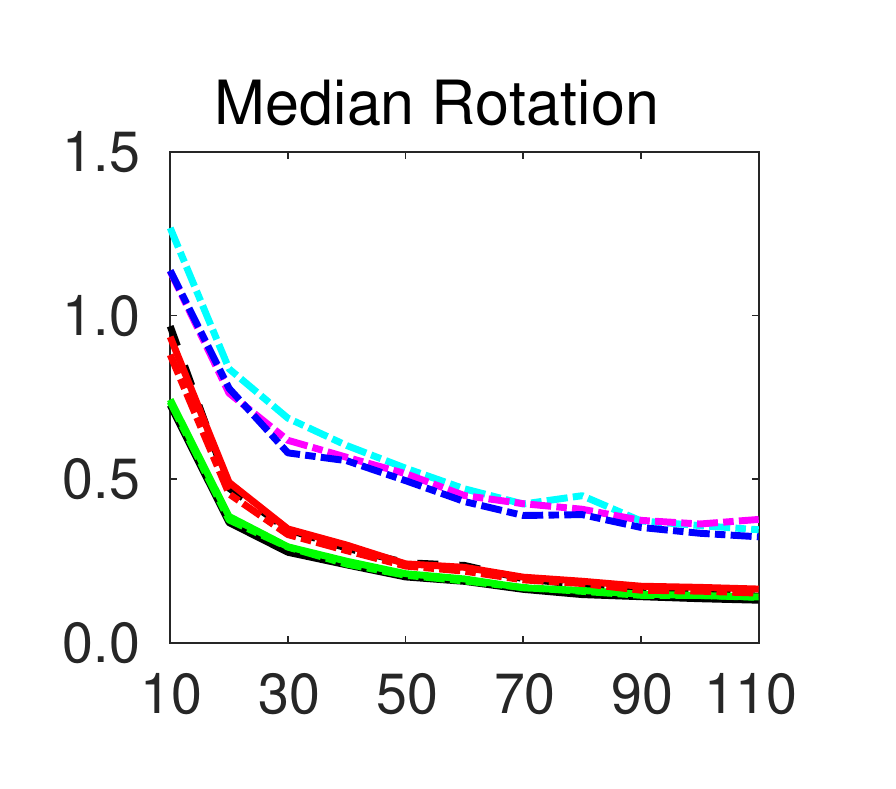}
\hspace{0cm}\includegraphics[trim={5mm 5mm 5mm 5mm},clip,width=0.24\textwidth,align=c]{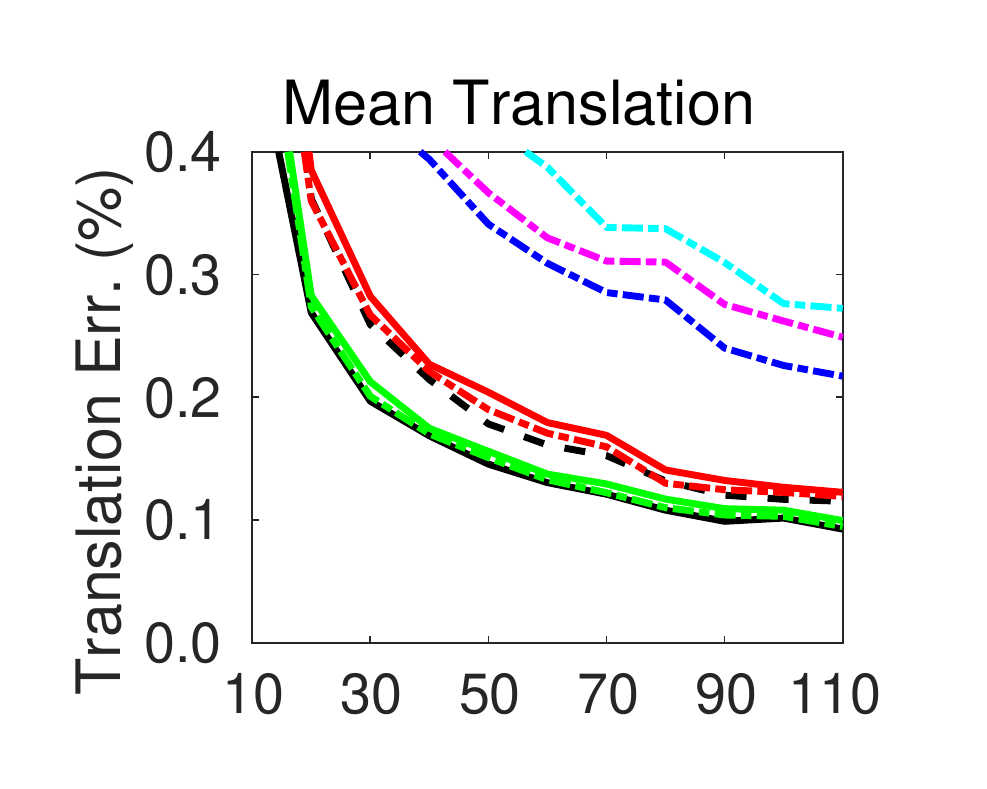} 
\hspace{0cm}\includegraphics[trim={5mm 5mm 5mm 5mm},clip,width=0.22\textwidth,align=c]{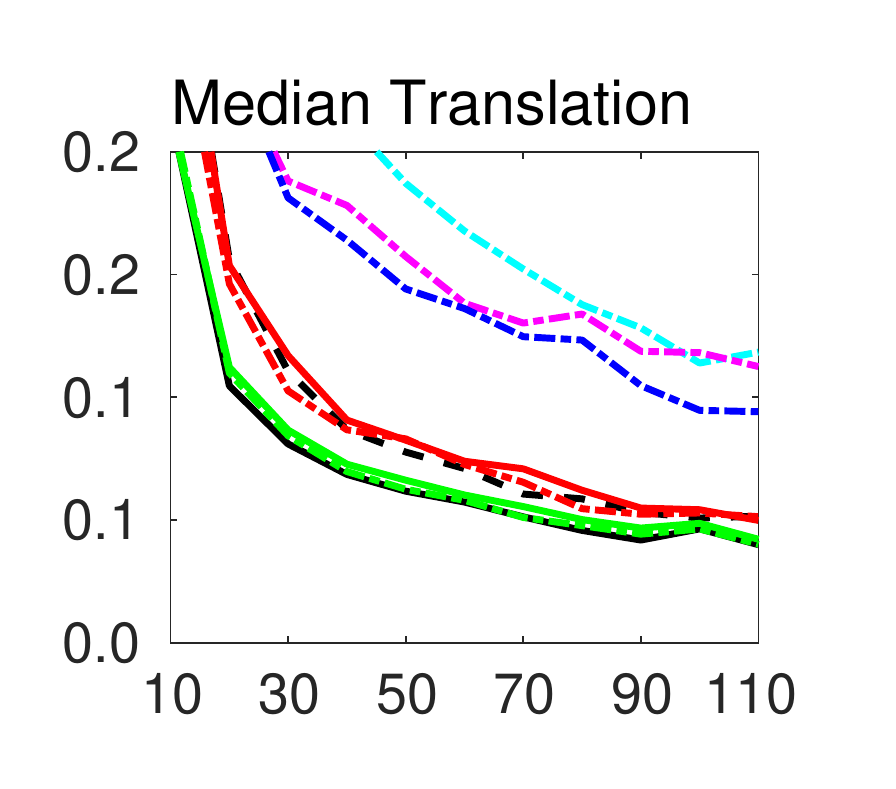}

\hspace{0cm}
\includegraphics[trim={5mm 5mm 88mm 15mm},clip,width=0.02\textwidth,align=c]{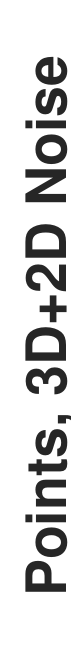} 
\includegraphics[trim={5mm 5mm 5mm 5mm},clip,width=0.24\textwidth,align=c]{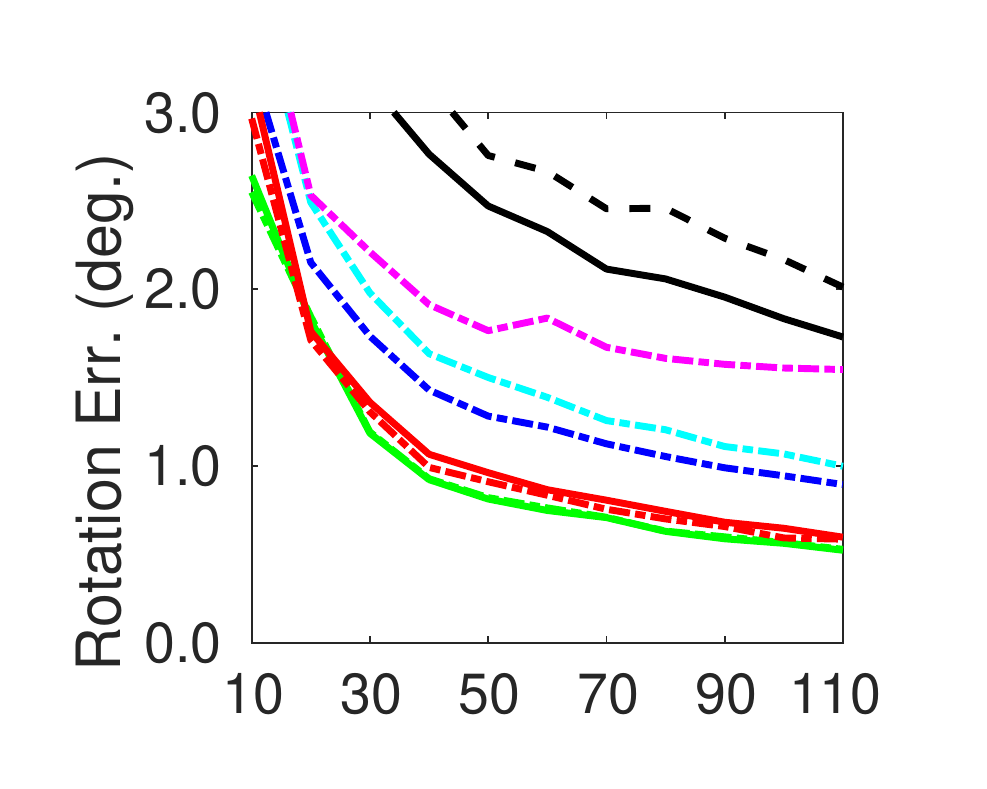} 
\hspace{0cm}\includegraphics[trim={5mm 5mm 5mm 5mm},clip,width=0.22\textwidth,align=c]{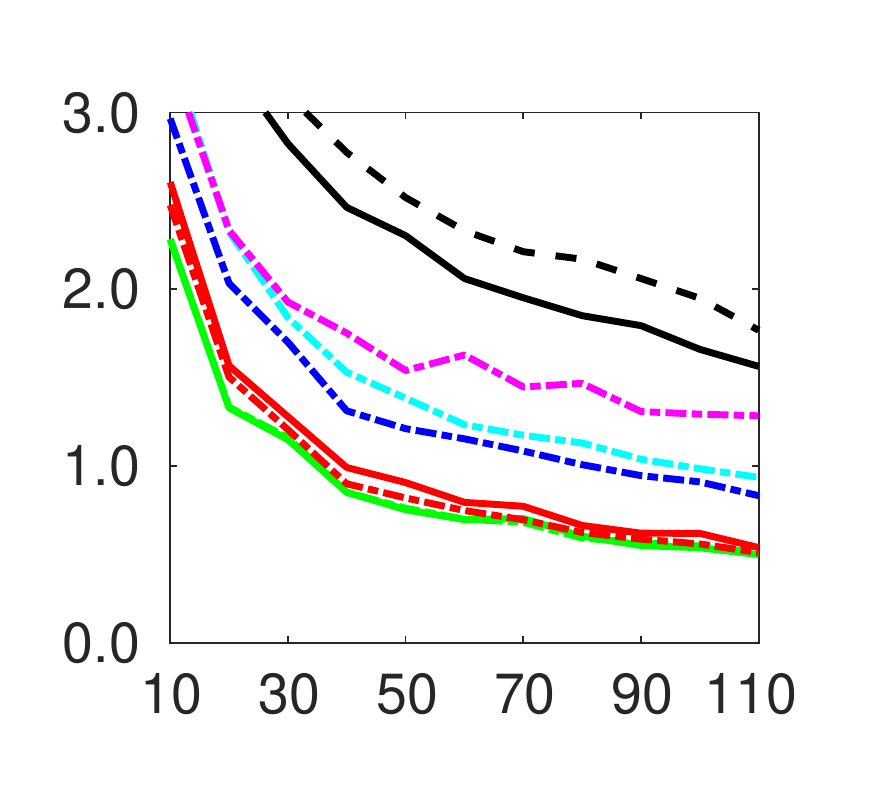}
\hspace{0cm}\includegraphics[trim={5mm 5mm 5mm 5mm},clip,width=0.24\textwidth,align=c]{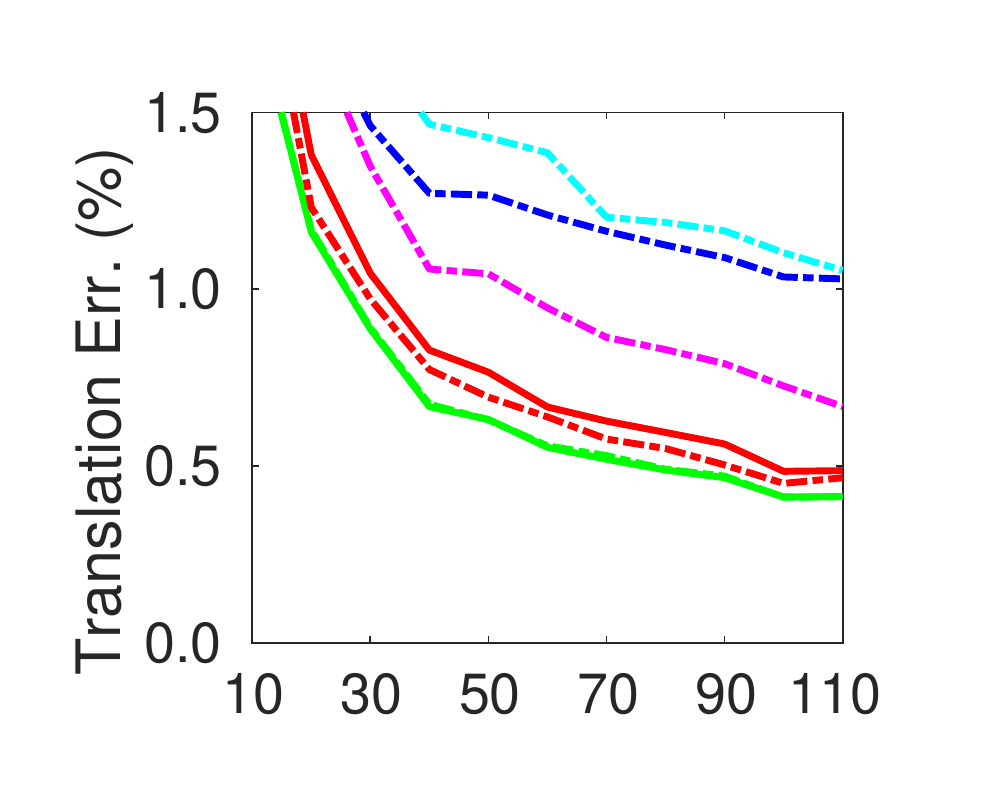} 
\hspace{0cm}\includegraphics[trim={5mm 5mm 5mm 5mm},clip,width=0.22\textwidth,align=c]{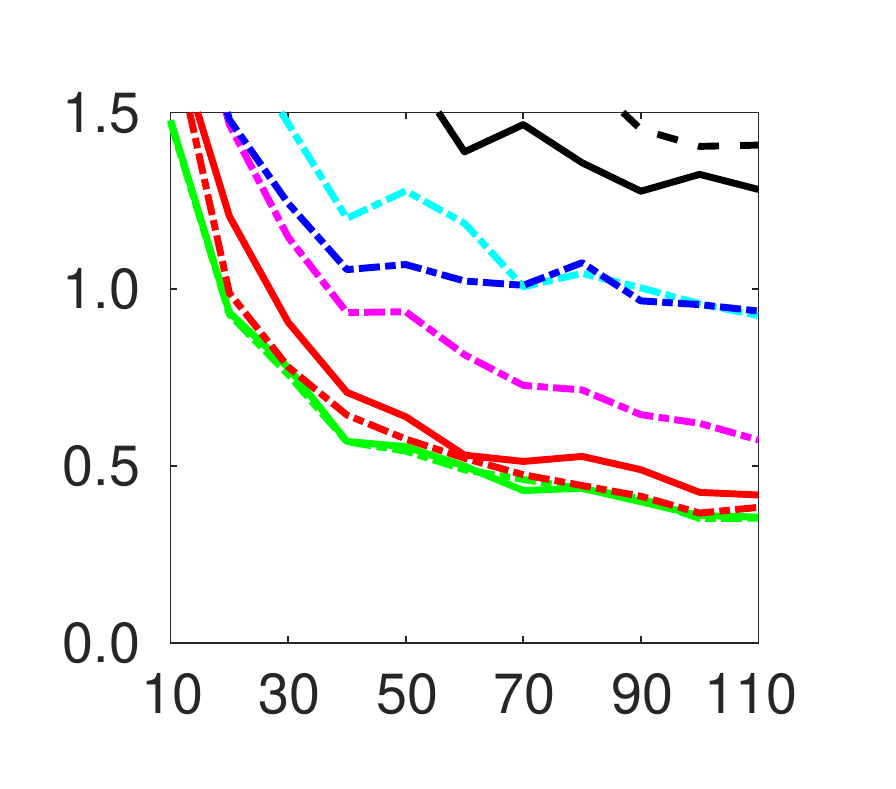}

\hspace{0cm}
\includegraphics[trim={5mm 0mm 88mm 10mm},clip,width=0.02\textwidth,align=c]{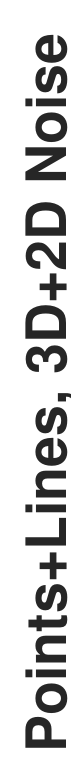} 
\includegraphics[trim={5mm 5mm 5mm 5mm},clip,width=0.24\textwidth,align=c]{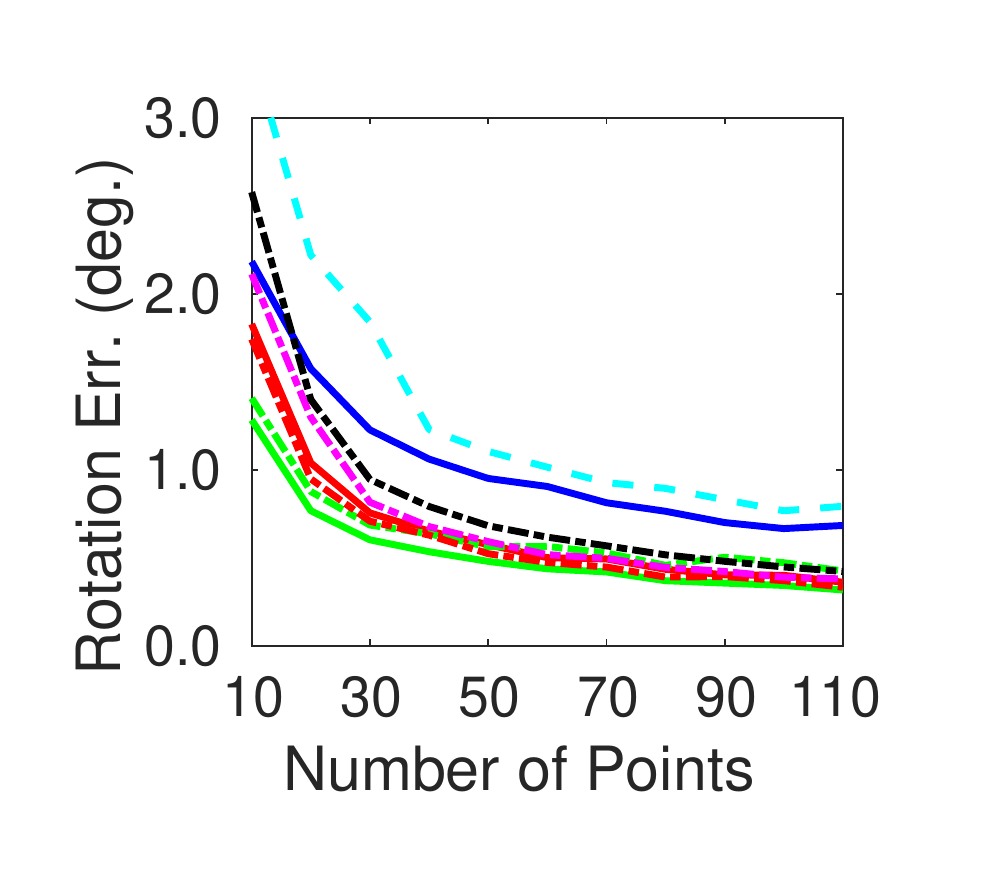} 
\hspace{0cm}\includegraphics[trim={5mm 5mm 5mm 5mm},clip,width=0.22\textwidth,align=c]{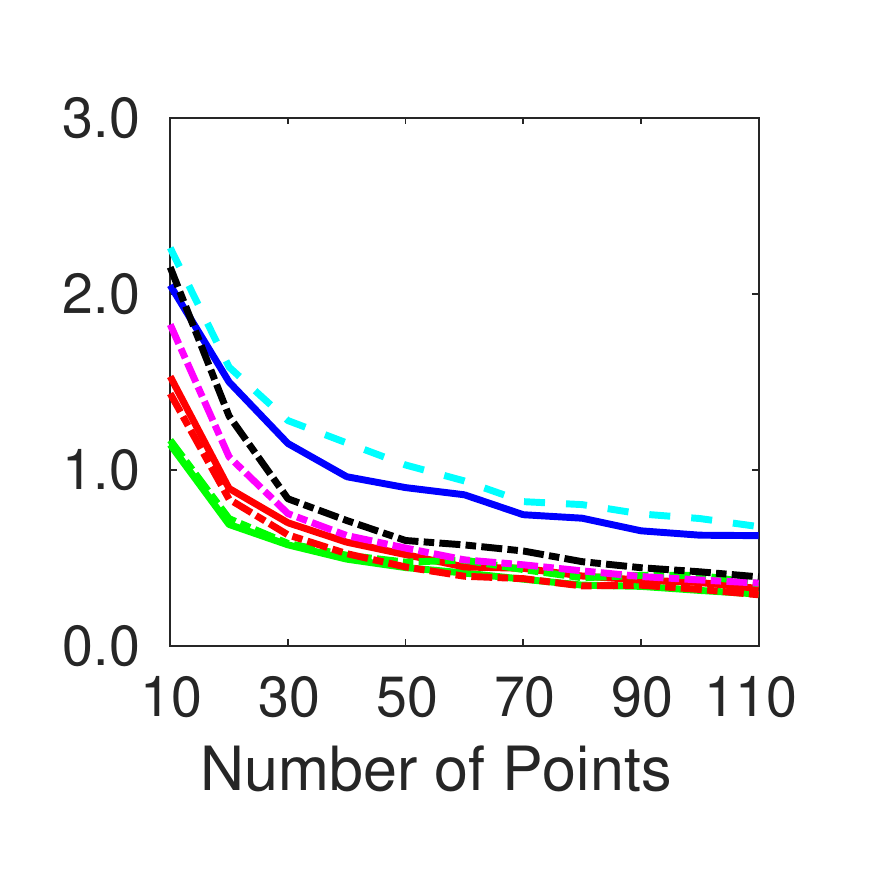}
\hspace{0cm}\includegraphics[trim={5mm 5mm 5mm 5mm},clip,width=0.24\textwidth,align=c]{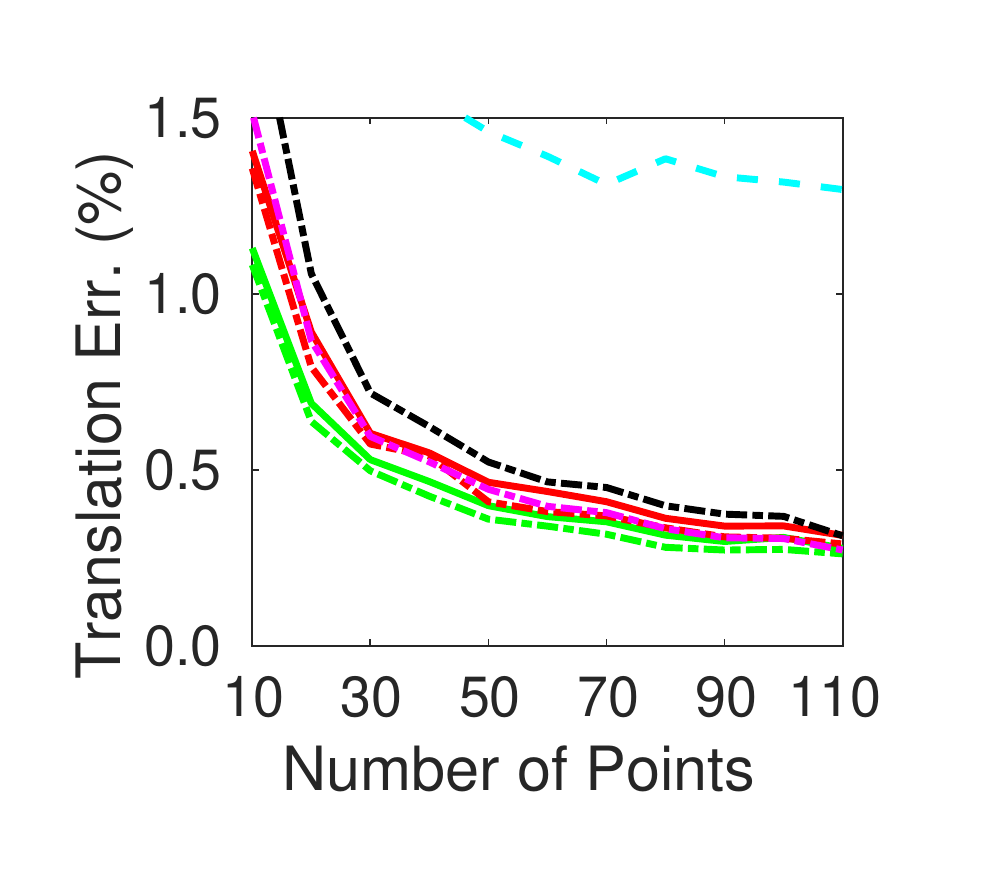} 
\hspace{0cm}\includegraphics[trim={5mm 5mm 5mm 5mm},clip,width=0.22\textwidth,align=c]{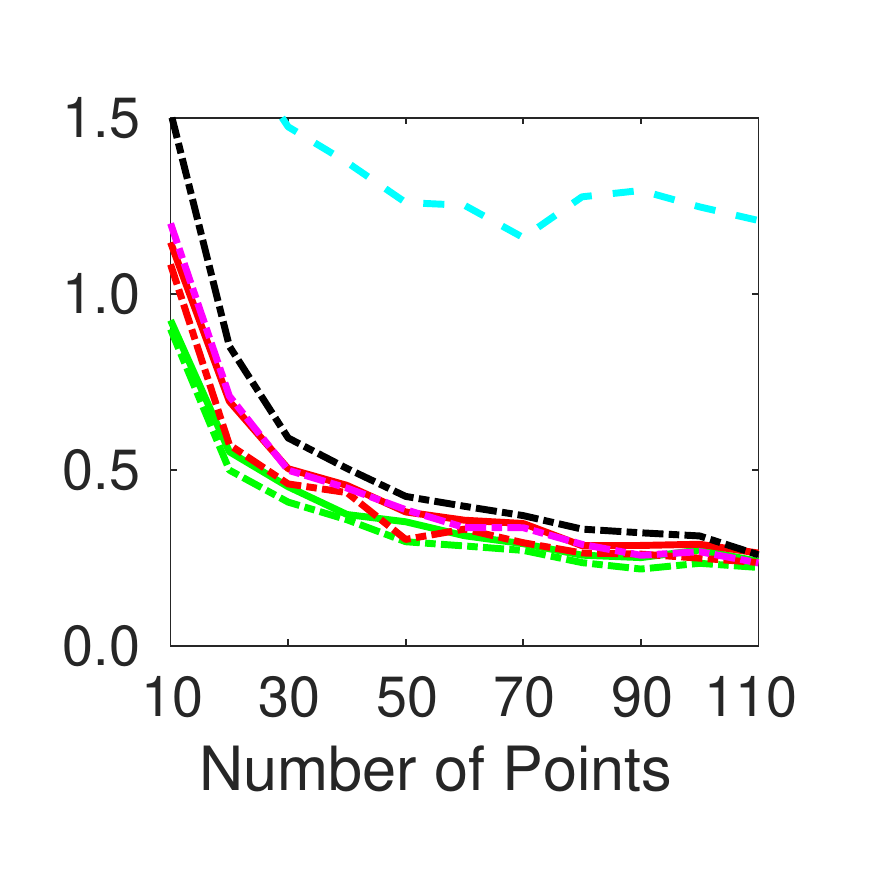}

\vspace{0cm}\includegraphics[trim={0mm 5mm 0mm 5mm},clip,width=0.55\textwidth]{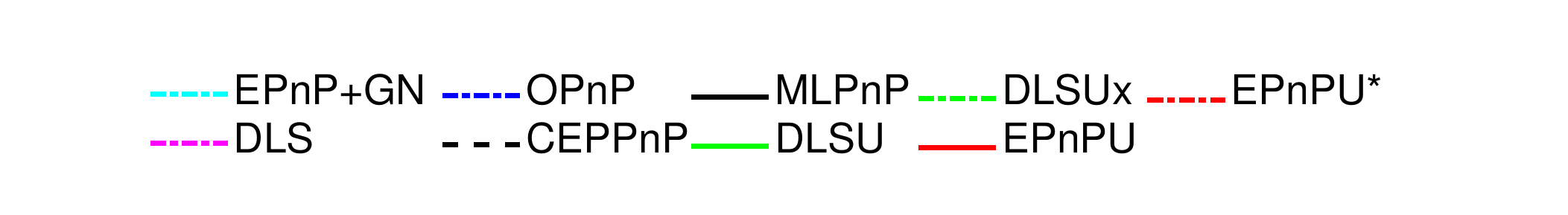}
\hspace{0cm}
\includegraphics[trim={0mm 5mm 0mm 5mm},clip,width=0.4\textwidth]{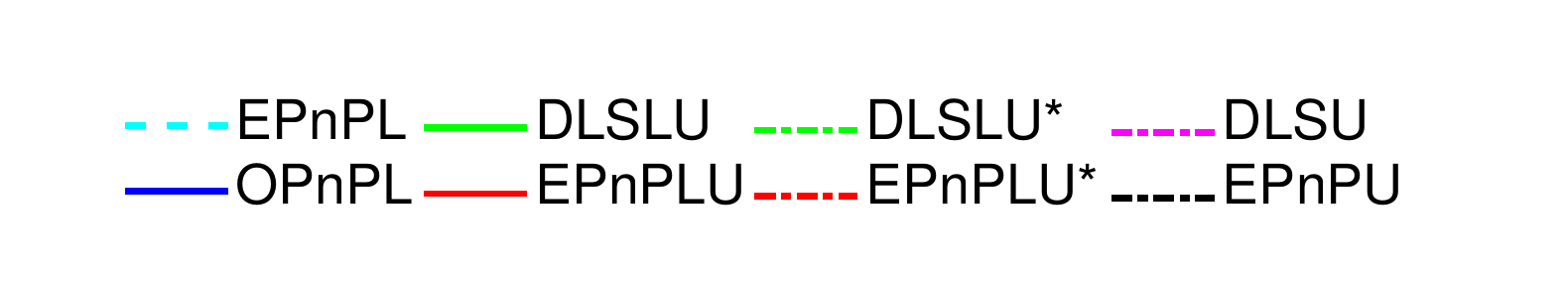}

\caption{Pose errors in synthetic experiments. Point-based pose estimation, increasing the number of points in case of 2D noise ({\bf Top}) and 2D+3D noise ({\bf Center}), left legend. {\bf Bottom:} increasing the number of lines and points in case of 2D+3D noise, right legend. We report mean and median rotation and translation errors. Asterisk denotes rough pose hypothesis  input.
In case of 2D noise the new approaches reach the state-of-the-art accuracy, in case of 3D noise they outperform the published methods.
In case of the line features, the new solvers outperform the  published EP$n$PL and OP$n$PL as well as the proposed uncertainty aware point-only methods. The access to a pose hypothesis does not result in better accuracy.
}\label{fig:synthetic1}
\vspace{-0.5cm}
\end{figure*}

\section{Experiments}\label{sec:exp}
We compare the proposed uncertainty-aware pose solvers to competitive pose estimation methods, in isolation and combined with a {\em standard} or {\em uncertain} refinement, see Section~\ref{sec:ref}. We use asterisk $*$ to denote the methods receiving pose hypothesis. We compare against EP$n$P~\cite{lepetit2009}, DLS~\cite{hesch2011direct},
2D covariance-aware methods CEPP$n$P~\cite{ferraz2014leveraging} and  MLP$n$P~\cite{isprs-annals-III-3-131-2016}, the state-of-the-art P$n$P method  OP$n$P~\cite{zheng2013} in case of points, and EP$n$PL and OP$n$PL~\cite{vakhitov2016accurate} in case of points and lines mixture.

We use RANSAC~\cite{fischler1981} with P3P~\cite{kneip2011novel} to estimate inliers before feeding them into the solvers, while also comparing against the P3P baseline which does not use any P$n$P pose solver. After the solvers, we optionally run inlier filtering using the obtained pose, followed by a motion-only bundle adjustment, inspired by the localization modules of ORB-SLAM2~\cite{mur2017orb} or COLMAP~\cite{schonberger2016structure}. We use MATLAB implementations of the methods, run our experiments on a laptop with Core i7 1.3 GHz with  16Gb RAM. 

\subsection{Synthetic experiments}

In the synthetic setting we compare the proposed pose estimation methods EP$n$P(L)U and DLS(L)U against the baselines in isolation, as well as the proposed {\em uncertain} refinement against the standard refinement, as defined in Section~\ref{sec:ref}. 

{\bf Metrics.} We evaluate the results in terms of the absolute rotation error $\erot =  |\textrm{acos}
(
0.5(
\textrm{trace}
(\mathtt{R}_{\textrm{true}}^T \mathtt{R})-1))|$ in degrees and relative translation error $\etrel = \lVert \bt_{\textrm{true}} - \bt \rVert / \lVert \bt_{\textrm{true}} \rVert  \times 100$, in $\%$, where $\mathtt{R}_{\textrm{true}},$ $\bt_{\textrm{true}}$ is the true pose and $\mathtt{R}$, $\bt$ is the estimated one. 

\newcommand{\STAB}[1]{\begin{tabular}{@{}c@{}}#1\end{tabular}}

\setlength\tabcolsep{3pt}
{\small 
\begin{table*}[t]
\small \small
\centering
\begin{threeparttable}
\centering
\begin{tabular}[t]{|c|c|c|c|c|c|c|c|c|c|c|c|c|c|c|c|c|c|c|c|c|}
\hline
 & \multicolumn{8}{|c|}{Points}
 & \multicolumn{4}{|c|}{Points + 2D Uncertainty}
  & \multicolumn{8}{|c|}{Points + Full Uncertainty, Proposed} \\
  \hline

 & \multicolumn{2}{ |c| }{ P3P~\cite{kneip2011novel} } & \multicolumn{2}{ |c| }{ EP$n$P~\cite{lepetit2009}} & \multicolumn{2}{ |c| }{ DLS~\cite{hesch2011direct} } & \multicolumn{2}{ |c| }{ OP$n$P~\cite{zheng2013} } &
 
 \multicolumn{2}{ |c| }{ CEPP$n$P~\cite{ferraz2014leveraging} } & \multicolumn{2}{ |c| }{ MLP$n$P~\cite{isprs-annals-III-3-131-2016}} &
 
 \multicolumn{2}{ |c| }{ EP$n$PU* } & \multicolumn{2}{ |c| }{ EP$n$PU } & \multicolumn{2}{ |c| }{ DLSU* } & \multicolumn{2}{ |c| }{ DLSU }
 \\
 \hline
  & $\erot$ & $\etrel$ 
 & $\erot$ & $\etrel$ 
 & $\erot$ & $\etrel$ 
 & $\erot$ & $\etrel$ 
 
 & $\erot$ & $\etrel$ 
 & $\erot$ & $\etrel$ 
 
 & $\erot$ & $\etrel$ 
 & $\erot$ & $\etrel$ 
 & $\erot$ & $\etrel$ 
 & $\erot$ & $\etrel$  \\
\hline

 \multicolumn{21} {|c|}{  KITTI~\cite{geiger2012we}, sequences 00-02 }\\
 \hline
 
 N & 8.6 & 35.2 & 4.5 & 24.0 & 5.5 & 18.1 & 7.8 & 277.6 & 8.2 & 49.5 & 5.8 & 27.2 & {\bf 4.2} & 22.2 & 5.1 & 23.9 & 5.6 & 32.2 & 6.0 & {\bf 14.9} \\ 
 S & 5.1 & 14.4 & 4.0 & 12.8 & 5.0 & 12.2 & 7.2 & 242.2 & 5.3 & 20.6 & 5.3 & 14.4 & {\bf 3.7} & 12.6 & 3.9 & 13.1 & 5.1 & 25.5 & 5.1 & {\bf 12.1} \\ 
 U & 5.0 & 14.0 & 3.5 & 13.2 & 5.0 & 12.9 & 7.6 & 325.5 & 5.1 & 17.4 & 6.3 & 35.3 & {\bf {\em 3.3}} & {\bf {\em 10.6}} & 3.5 & 13.4 & 5.0 & 12.6 & 5.0 & 10.9 \\ 
 
 \hline

 \hline
 \multicolumn{21} {|c|}{ TUM~\cite{sturm12iros}, 'freiburg1' sequences} \\
 \hline

 N & 15.7 & 3.3 & 9.5 & 1.5 & 9.3 & 1.4 & 10.0 & 1.5 & 10.0 & 1.7 & 10.0 & 1.6 & 9.3 & 1.3 & 9.4 & 1.4 & 9.3 & 1.3 & {\bf 9.1} & {\bf 1.2} \\ 
 S & 9.2 & 1.2 & 9.0 & 1.2 & 9.0 & 1.2 & 9.7 & 1.3 & 9.4 & 1.3 & 9.2 & 1.2 & 9.0 & 1.2 & 9.0 & 1.2 & 9.0 & 1.2 & {\bf 9.0} & {\bf 1.2} \\ 
U & 9.1 & 1.2 & 9.0 & 1.2 & 9.0 & 1.2 & 10.3 & 1.3 & 9.2 & 1.2 & 9.6 & 2.0 & {\bf {\em 9.0}} & 1.1 & 9.0 & 1.2 & 9.0 & {\bf {\em 1.1}} & 9.0 & 1.1 \\ 

\hline

\end{tabular}
\caption{Motion estimation from 2D-3D point correspondences on KITTI~\cite{geiger2012we}  TUM~\cite{sturm12iros} in terms of mean absolute rotation $\erot$  (in $0.1 \times $deg.) and translation $\etrel$  (in cm.) errors. We compare proposed full uncertainty-aware methods against point-based P$n$P and 2D uncertainty-aware methods in isolation ({\bf N}), with standard ({\bf S}) and proposed uncertain ({\bf U}) refinement. Methods with '*' receive a pose from RANSAC, best for the dataset is in bold italic, best for each protocol (N,S or U) is in bold. The new methods outperform the baselines in most metrics,e.g. DLSU in isolation improves $\etrel$ on KITTI {\bf by 3 cm (18\%)} compared to the best performing baseline DLS. {\em Uncertain} (U) is mostly better than standard (S) for the proposed methods, e.g.  $\etrel$  {\bf by 2 cm (16\%)} for EP$n$PU* on KITTI.}
\label{table:accuracy}

\end{threeparttable} 
\vspace{-0.5cm}
\end{table*}
}
{\bf Data generation.} We assume a virtual calibrated camera with an image size of $640\times480$ pix., a focal length of 800 and a principal point in the image center. 3D points and endpoints of 3D line segments are generated in the box \mbox{$[-2, 2] \times [-2, 2] \times [4, 8]$} defined in camera coordinates. 3D-to-2D correspondences are then defined by projecting the 3D points under the random rotation matrix and translation vector. We move the 3D line endpoints randomly along the line by a randomly generated Gaussian shift with a standard deviation equal to 10$\%$ of the 3D line length, see~\cite{vakhitov2016accurate}.

We add noise of varying magnitude to the 2D point or line endpoint projections, as well as to the 3D points or line endpoints,

 splitting $n_{pt}$ points into 10 subsets with an equal number of points in order to introduce differences in the noise magnitude. Each subset is corrupted by Gaussian noise with an increasing value of standard deviation, from $\sigma=0.05$ to $\sigma=0.5$. We consider anisotropic covariances, which are computed by randomly picking a rotation and a triplet $\{\sigma,\sigma_1,\sigma_2\}$, where $\sigma_1,\sigma_2$ are random values chosen within the interval $(0,\sigma]$. The covariance axes are scaled and rotated according to a triplet of standard deviations and the rotation value, respectively. We perform exactly the same addition of noise to the 3D endpoints of line segments. We add noise with different variance to the point and line endpoint projections using the same mechanism increasing the standard deviation from $\sigma=1$ to $\sigma=10$. 

We perform 400 simulation trials. The experiment settings are consistent with~\cite{zheng2013,ferraz2014very,ferraz2014leveraging,vakhitov2016accurate}.
We evaluate the pose solvers in isolation in a point-only and a point+line setting, providing the methods marked by '*' a pose hypothesis computed from a randomly chosen subset of three points using P3P~\cite{kneip2011novel}. We change $n_{pt}=10$ to $110$, in two different setups: introducing noise to the projected 2D features and introducing noise also to the 3D points or endpoints. In the case of experiments with lines and points, we generate $n_l=n_{pt}$ line correspondences in addition to points. 

\begin{figure}[]
\centering 
\hspace{0.0cm}\includegraphics[trim={5mm 5mm 5mm 5mm},clip,width=0.22\textwidth]{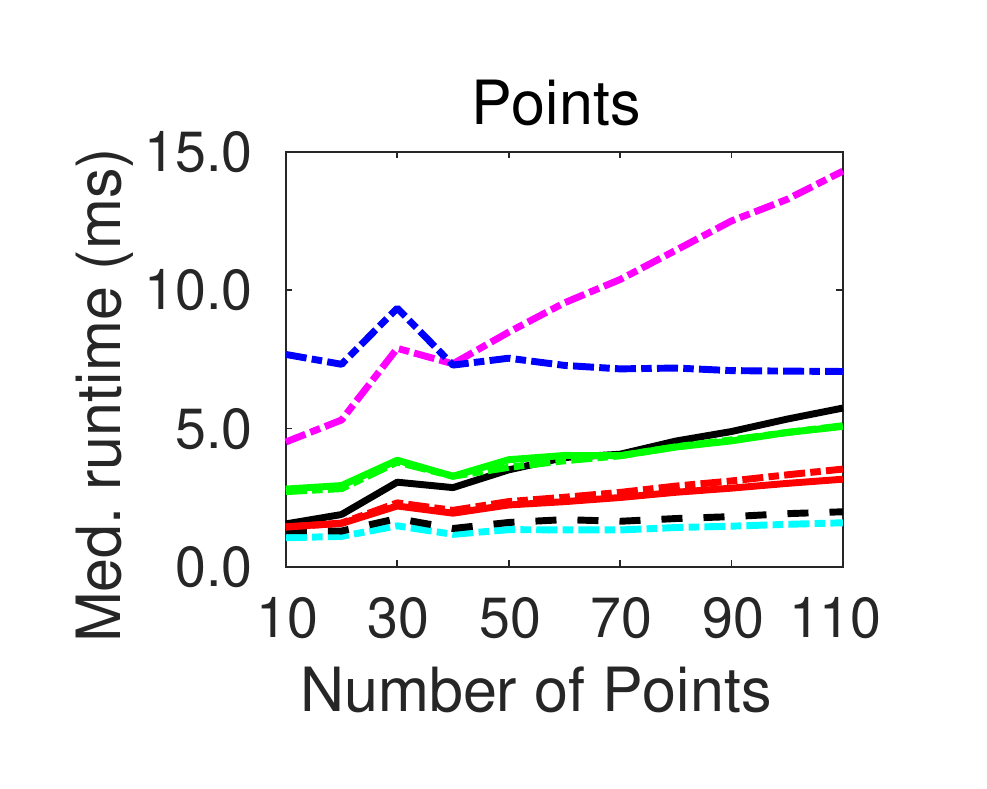} 
\hspace{0.5cm}\includegraphics[trim={5mm 5mm 5mm 5mm},clip,width=0.22\textwidth]{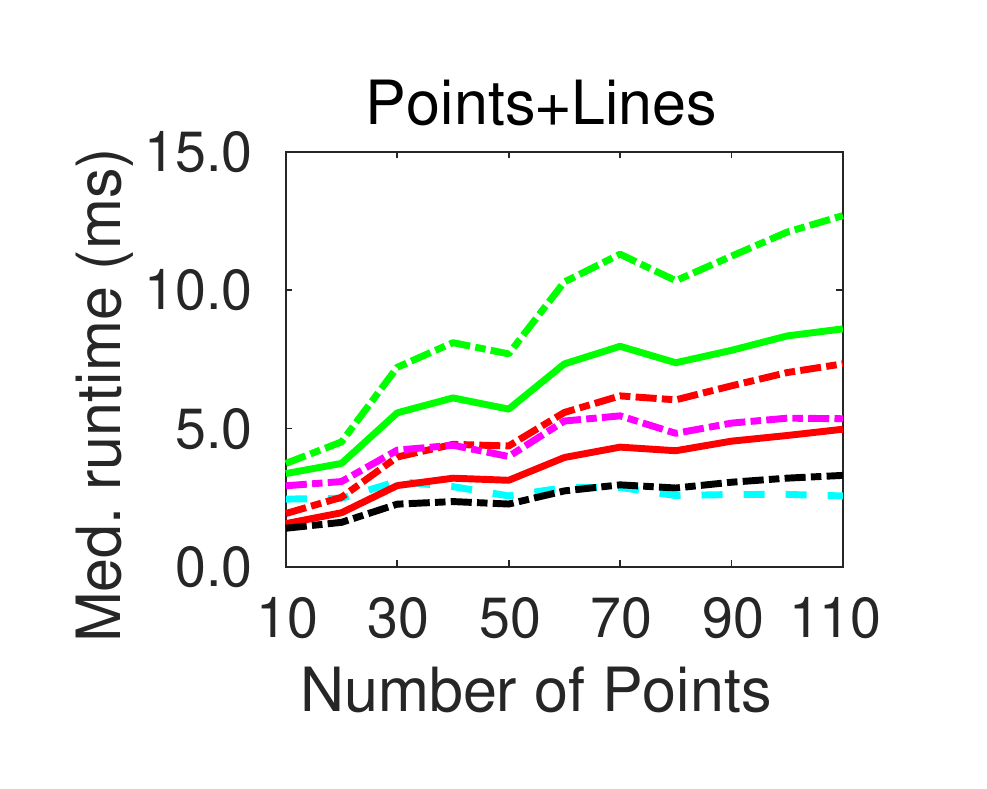}\\
  \caption{Runtime (ms). Methods based on points (left) or points and lines (right). See Fig.~\ref{fig:synthetic1} for the legends.}
  \label{fig:syntheticcost}
\vspace{-0.3cm}  
\end{figure}

{\bf Results.} Fig.~\ref{fig:synthetic1} summarizes the results of the experiments. In the 2D noise experiment for points (top row), the proposed methods perform similarly with 2D methods, however for $n_{pt}<30$ the MLP$n$P delivers slightly better results, probably due to additional reprojection error refinement step used in this method and not used in the other ones. When we use both 3D and 2D noise for the points (central row), the proposed methods are the most accurate, followed by the classical P$n$P solvers, and the 2D covariance-based methods. In point+line experiment, the new methods clearly outperform the baselines.  Fig.~\ref{fig:syntheticcost} shows an analysis in terms of computational cost. The fastest are EP$n$PU, EP$n$P, CEPP$n$P, MLP$n$P, followed by DLSU, DLS  and OP$n$P. 

\begin{figure}[]
\centering 
\hspace{0.0cm}\includegraphics[trim={5mm 5mm 5mm 5mm},clip,width=0.22\textwidth]{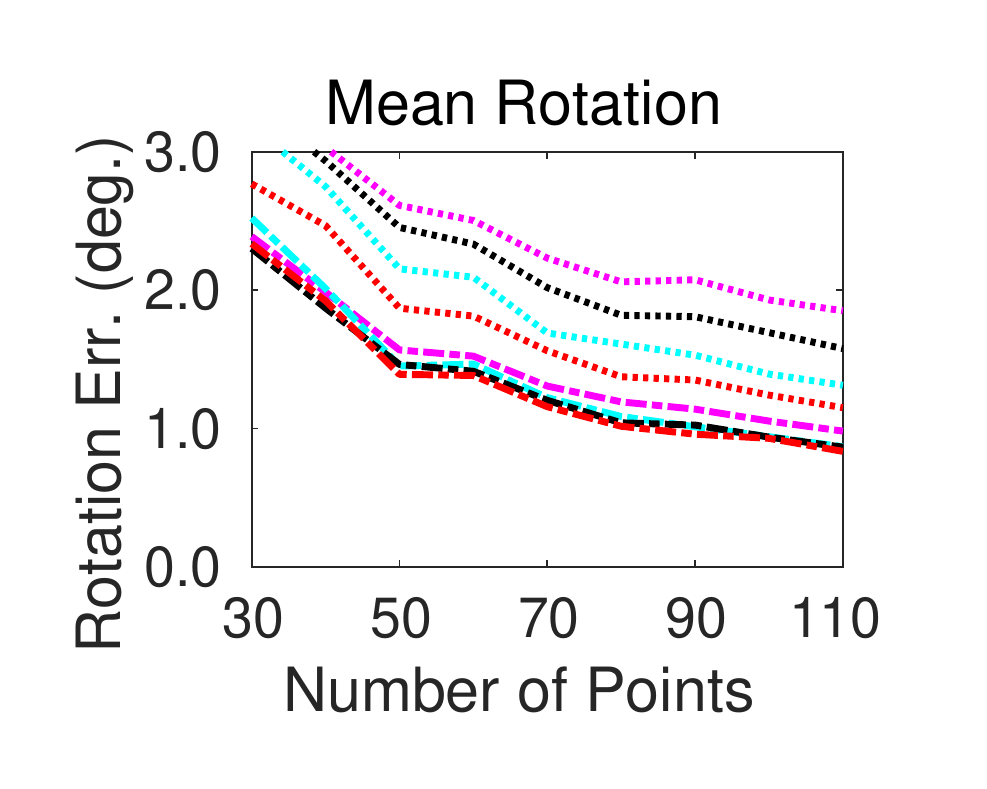} 
\hspace{0.5cm}\includegraphics[trim={5mm 5mm 5mm 5mm},clip,width=0.22\textwidth]{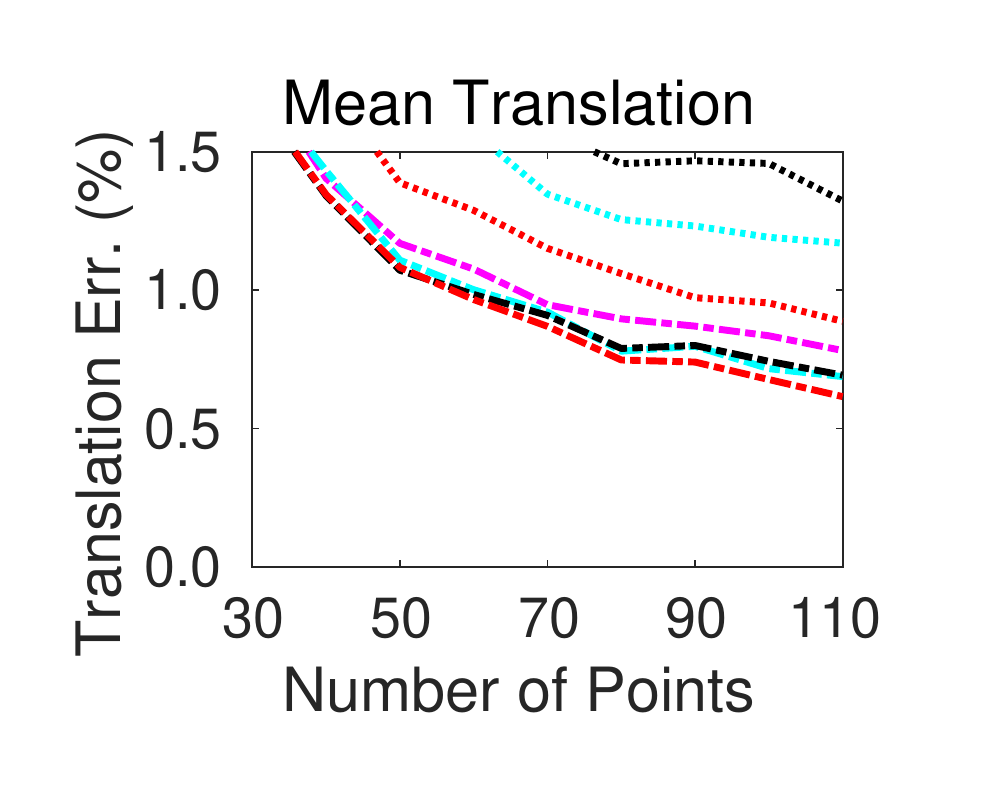}\\

\vspace{0cm}\includegraphics[trim={0mm 5mm 0mm 5mm},clip,width=0.45\textwidth]{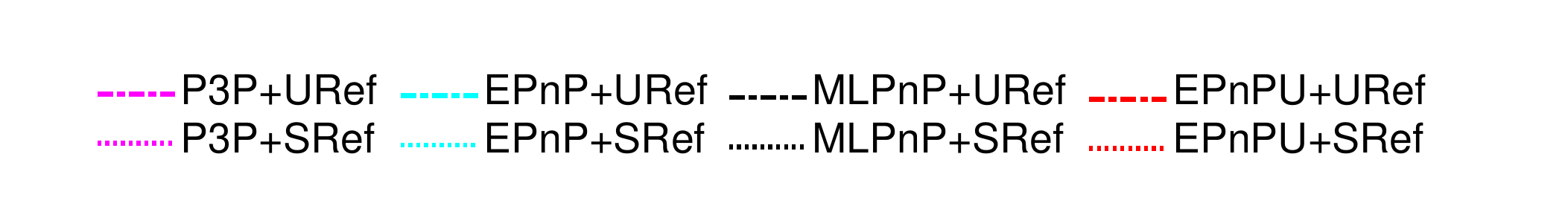}
\hspace{0cm}
  \caption{Comparison of methods with standard(+SRef) or uncertain(+URef) refinement, in a 2D + 3D noise setup as in Fig.~\ref{fig:synthetic1}, central row. Uncertain refinement improves accuracy, the uncertainty-aware EP$n$PU* is slightly better than the other methods.}
  \label{fig:refinement}
\vspace{-0.4cm}  
\end{figure}

In Fig.~\ref{fig:refinement}, we compare the proposed {\em uncertain} refinement against the {\em standard} method for point features, see Section~\ref{sec:ref}. For the inlier filtering, we use a threshold $\tau^2=6^2$ for the covariance-weighted squared residuals~(\ref{eq:ransac_pt_residual}) corresponding to the {\em standard} or the {\em uncertain} refinement. The data is generated as in the experiment with 3D and 2D noise. We consider EP$n$P, MLP$n$P and EP$n$PU*. The {\em uncertain} refinement is beneficial for all considered solvers; the margin between different pose solvers after refinement decreases, but remains, because the more accurate pose solvers can provide a better set of inliers for the final refinement step; see additional results on timing and comparison against the {\em full uncertain} refinement in the supp. mat.

Summarizing, the new methods outperform baselines in a synthetic setting. In the next section, we show that the same holds for the real scenarios.

\subsection{Real experiments}
\begin{table}[t]
\centering
\begin{threeparttable}
\centering
\begin{tabular}[t]{
|c|c|c|c|
c|c|c|c
|c|}
\hline
 & \multicolumn{4}{|c|}{Points+Lines}
 & \multicolumn{4}{|c|}{Points+Lines+Uncertainty}
  \\
  \hline
& \multicolumn{2}{ |c| }{ EPnPL~\cite{vakhitov2016accurate} } & \multicolumn{2}{ |c| }{ OPnPL~\cite{vakhitov2016accurate} }  & \multicolumn{2}{ |c| }{ DLSLU* } &  \multicolumn{2}{ |c| }{ EPnPLU* } \\ 
\hline
  & $\erot$ & $\etrel$ 
 & $\erot$ & $\etrel$ 
 & $\erot$ & $\etrel$ 
 & $\erot$ & $\etrel$ \\
 \hline
 N & {\bf 2.5} & 37.1 & 10.2 & 650.1 & 6.3 & {\bf 18.2} &
   3.4 & 25.2  \\ 
 S & 1.8 & 20.4 & 6.8 & 267.4 
 & 5.2 & 12.2  & {\bf {\em 1.8}} & {\bf {\em 9.8}}  \\ 
 U & 1.4 & 12.1 & 9.0 & 497.7 & 5.2 & 12.0 & {\bf {\em 1.4}} & {\bf {\em 9.3}} \\
 \hline
\end{tabular}

\caption{Motion estimation from 2D-3D point and line correspondences on KITTI~\cite{geiger2012we} sequences 00-02. We report the mean rotation errors in $0.1 \times $degrees and translation errors in cm, for the solvers in isolation (N), after {\em standard} (S) and uncertain (U) refinement, see Section~\ref{sec:ref}. Proposed EP$n$PLU*, DLSLU* mostly outperform the  baselines OP$n$PL and EP$n$PL, e.g. $\etrel$ {\bf by 23\%-52\% (3 - 11 cm.)}, while EP$n$PL has the best rotation accuracy in isolation.}
\label{table:accuracy_lines}
        
\end{threeparttable} 
\vspace{-0.5cm}
\end{table}

{\small
\begin{table}[t]
\small \small
\centering
\begin{threeparttable}
\centering
\small
\begin{tabular}[t]{
|c|c|c|c|
c|c|c|c
|c|}
\hline
 & P3P  & EPnP  & DLS  & OPnP  &  \begin{tabular}{@{}c@{}}CEP- \\ PnP\end{tabular}  & \begin{tabular}{@{}c@{}}ML- \\ PnP\end{tabular}  & EPnPU   & DLSU \\ 
 \hline
 N &3.1&4.6&16.5&10.7&5.0&7.6&6.1&8.0 \\ 
 S &12.1&13.2&25.1&19.5&13.7&16.0&14.7&16.7 \\ 
U &11.3&12.7&24.5&18.8&13.1&15.2&13.9&16.0 \\ 
\hline
\end{tabular}
\caption{Average running time (ms) for the compared methods on KITTI in isolation (N), with standard (S) or uncertain (U) refinement. 
}
\label{table:runtime}
\end{threeparttable} 
\vspace{-0.5cm}
\end{table}
}

{\bf Data.} We use three monocular RGB sequences 00-02 from the KITTI dataset~\cite{geiger2012we} and the first three 'freiburg1' monocular RGB sequences of the TUM-RGBD dataset~\cite{sturm12iros} to evaluate the methods. We use KITTI in a monocular mode, taking a temporal window of two left frames ( three frames with a pose distance $>2.5cm$ for TUM) to detect and describe features, relying on FAST~\cite{rosten2006machine} and ORB~\cite{rublee2011orb} for points and EDLines~\cite{akinlar2011edlines} and LBD~\cite{zhang2013efficient} for lines, OpenCV implementations. We use an image pyramid with $n_o=8$ levels and a factor $\kappa=1.2$, the detection error $\epsilon=1$ pix, the uncertainty is calculated as described in Section~\ref{sec:cov}. The features are matched using standard brute-force approach, and triangulated using ground-truth camera poses.  Triangulation results are refined with Ceres~\cite{agarwal2012ceres}, producing also the 3D feature covariances, see supp. mat. for the detailed formulation.
The next left frame  in KITTI (the next RGB frame in TUM) is used for evaluation. 
 Line detections are filtered by length (less than 25 pix.~removed). We use a threshold of $\tau = 5.991$ for the covariance-weighted residual norms in   RANSAC. In case of point+line combination, we generate minimal sets using only points, include the lines into the motion-only bundle adjustment for the line-aware solvers.
 
 \begin{figure}[t]
	\begin{center}
    	\includegraphics[trim={5mm 10mm 88mm 15mm},clip,width=0.02\textwidth,align=c]{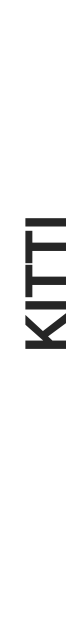} 
		\includegraphics[trim={5mm 5mm 5mm 5mm},clip,width=0.22\textwidth,align=c]{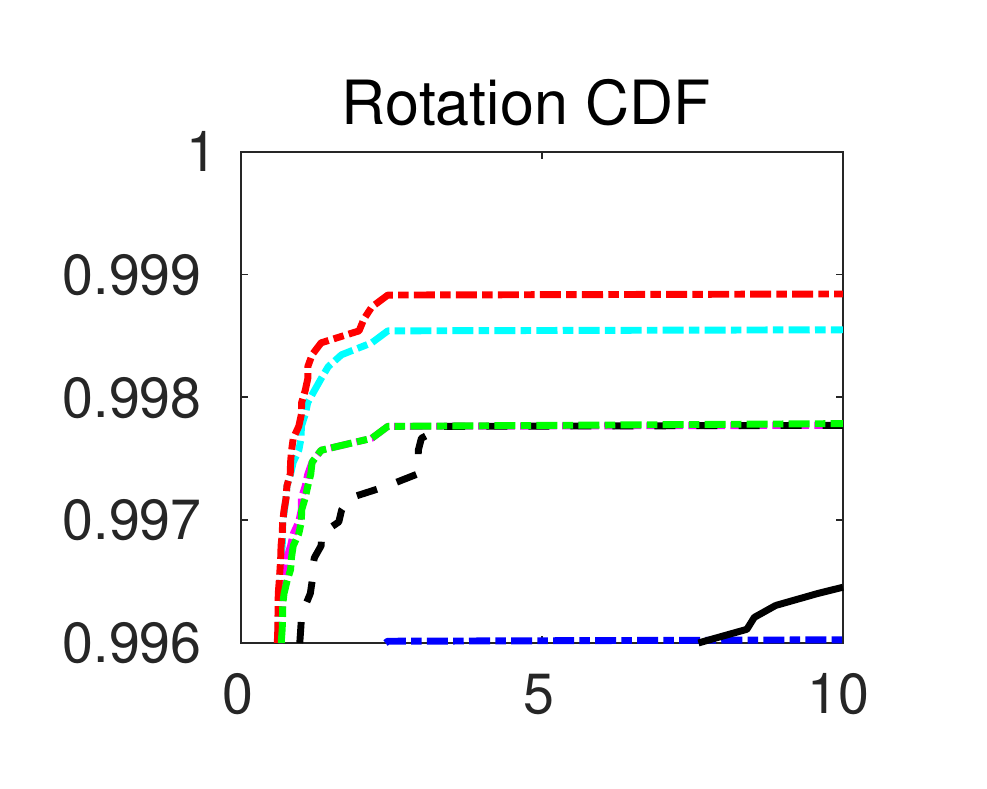}
		\includegraphics[trim={5mm 5mm 5mm 5mm},clip,width=0.22\textwidth,align=c]{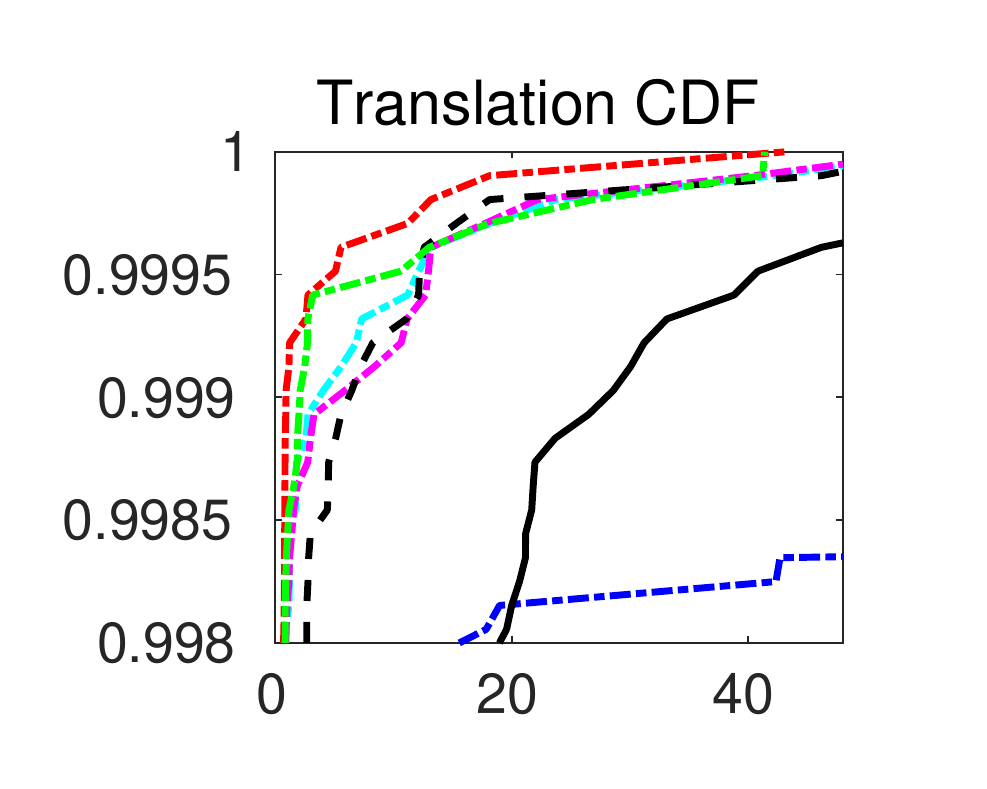}
		
		\includegraphics[trim={5mm 10mm 88mm 15mm},clip,width=0.02\textwidth,align=c]{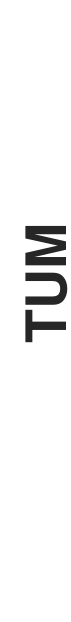} 
		\includegraphics[trim={5mm 5mm 5mm 5mm},clip,width=0.22\textwidth,align=c]{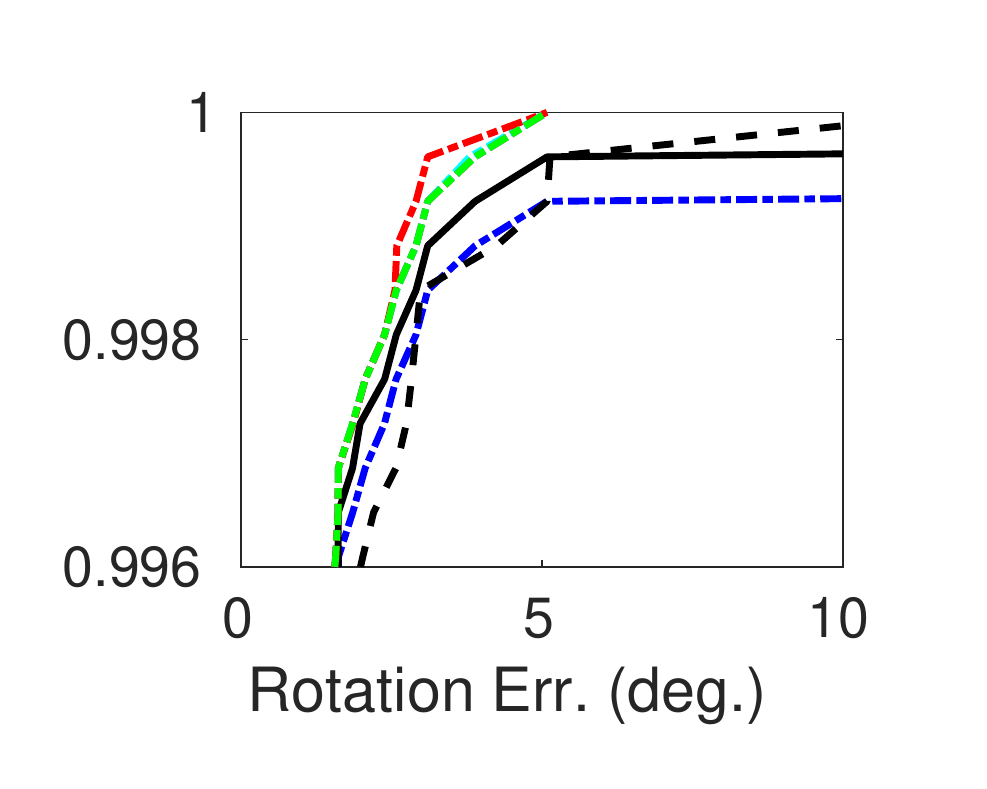}
		\includegraphics[trim={5mm 5mm 5mm 5mm},clip,width=0.22\textwidth,align=c]{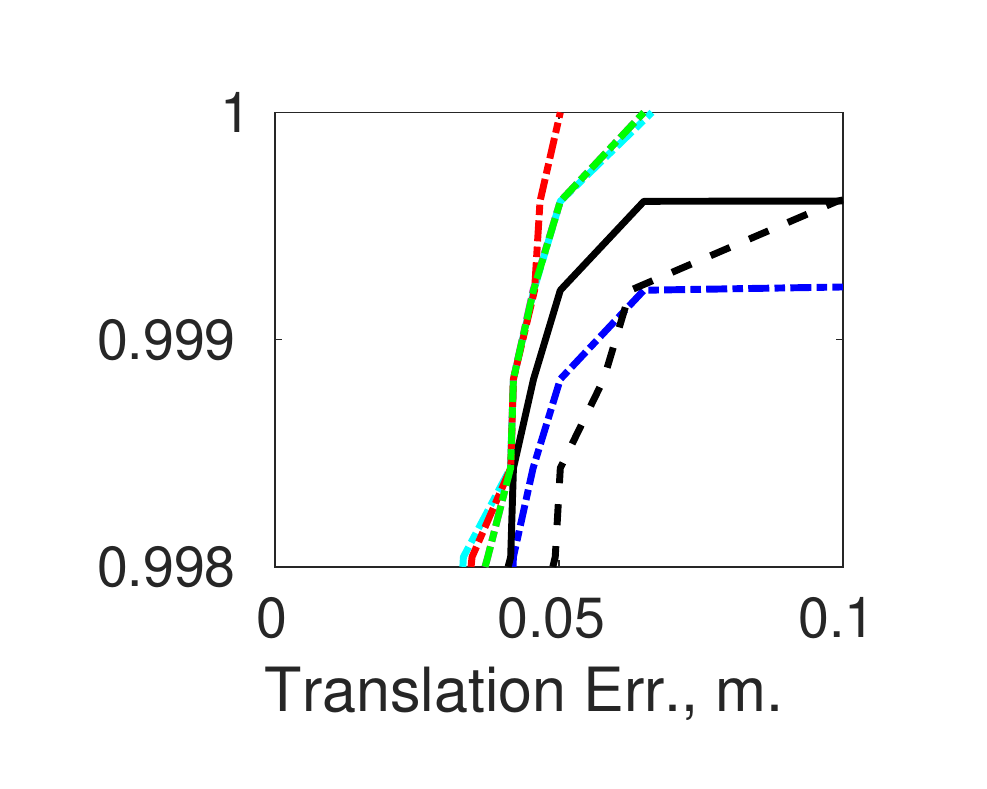}
		\includegraphics[trim={15mm 4mm 20mm 8mm},clip,height=8mm]{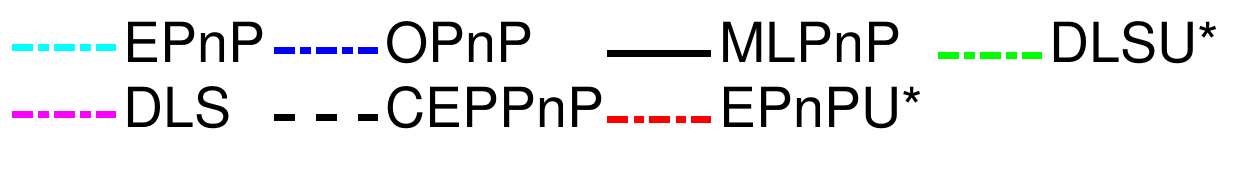}
	\end{center}
	\vspace{-0.3cm}
	\caption{CDF plots for real experiments on KITTI~\cite{geiger2012we} and TUM~\cite{sturm12iros}, U mode. EP$n$PU*  and DLSU* are the most accurate.}
	\label{fig:cdf}
\end{figure}
 
 \begin{figure}[]
\centering 
\hspace{0.0cm}\includegraphics[trim={5mm 110mm 5mm 110mm},clip,width=0.45\textwidth]{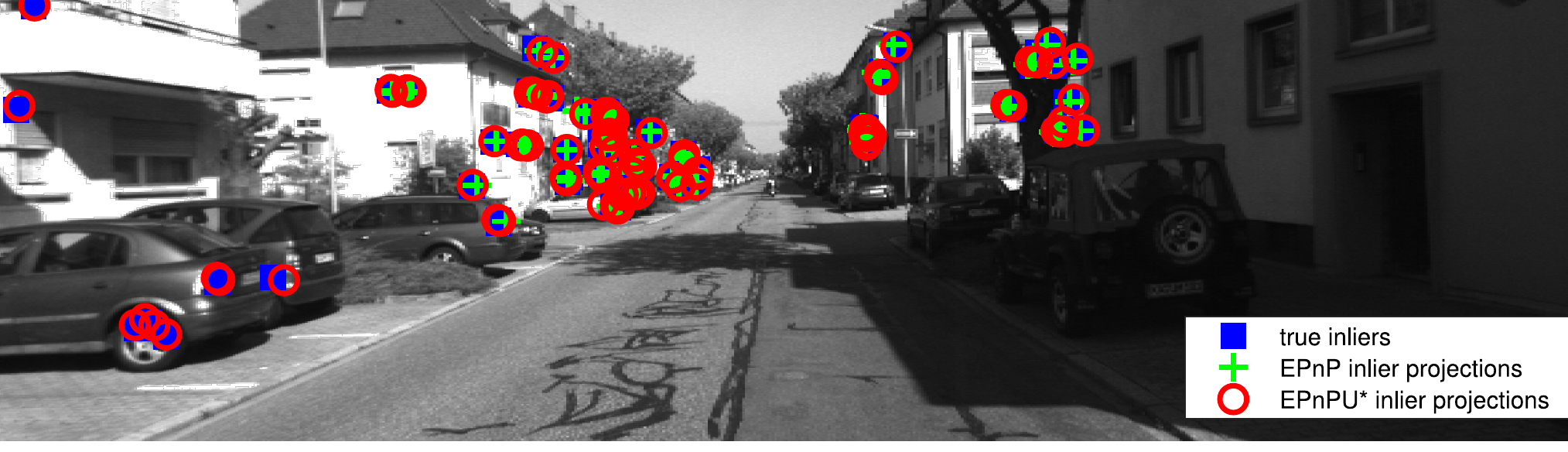} 
\hspace{0.5cm}
\includegraphics[trim={5mm 110mm 5mm 110mm},clip,width=0.45\textwidth]{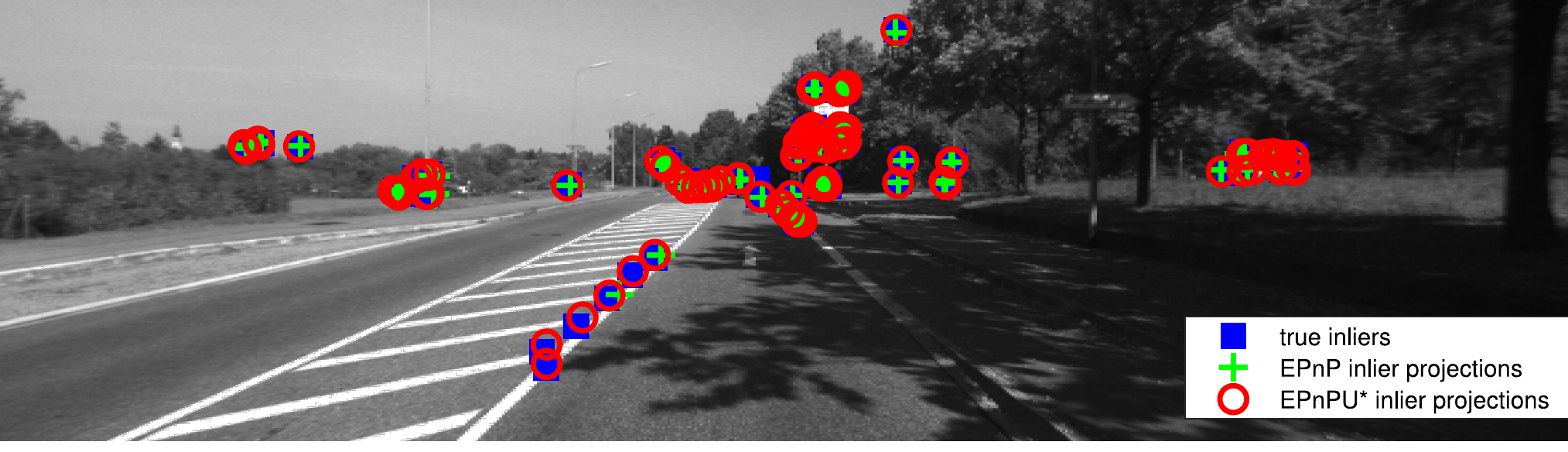} 
  \caption{
  Comparison of inlier sets from EP$n$PU*~(red) and EP$n$P~(green), blue squares show the true inliers. EP$n$PU* inlier sets are more complete.
  }
  \label{fig:examples}
\vspace{-0.5cm}  
\end{figure}
 
 {\bf Protocol and metrics.} We compare the methods using absolute rotation error in deg. and absolute translation error in cm. If a pose solver fails or produces less than 3 inliers, we do not use its output, but use the output of RANSAC instead, following~\cite{schonberger2016structure, mur2017orb}. 
 
 {\bf Results.} We evaluate the pose solvers in isolation, see Table~\ref{table:accuracy}, and show a significant increase in accuracy compared to the state-of-the-art,  e.g. DLSU on KITTI outperforms the closest baseline DLS by 18\% in mean translation.  On the TUM sequences, the mean rotation errors of the proposed methods are similar to the ones of the baselines, while there are more significant gains in translation errors. The proposed {\em uncertain} refinement improves accuracy over standard refinement for most of the solvers, e.g. by 16\% in case of EP$n$P on KITTI, see Table~\ref{table:accuracy} and CDF error plots in Fig.~\ref{fig:cdf}; the running time of the methods is in Table~\ref{table:runtime}. In the supp. mat. we give additional results, including median errors. See Fig.~\ref{fig:examples} for a visual comparison of the inlier sets estimated by EP$n$P and  EP$n$PU*. 
 
 In Table~\ref{table:accuracy_lines} we report the mean errors of the points-and-lines-based pose estimation for the proposed solvers. We observe an improvement in the translation errors by almost 50\% for the solvers in isolation and by 24\% after the standard refinement, compared to the uncertainty-free EP$n$PL and OP$n$PL~\cite{vakhitov2016accurate} methods.
 
\section{Conclusions}
We have generalized P$n$P(L) methods to estimate the camera pose with uncertain 2D feature detections and 3D feature locations and proposed a new pose refinement scheme. Our methods demonstrate increased accuracy and robustness both in synthetic and real experiments.